\documentclass[10pt,twocolumn,letterpaper]{article}

\usepackage{cvpr}              

\usepackage{adjustbox}
\usepackage{multirow}
\usepackage{tabularx}

\usepackage{amssymb}

\usepackage[accsupp]{axessibility} 

\definecolor{cvprblue}{rgb}{0.21,0.49,0.74}
\usepackage[pagebackref,breaklinks,colorlinks,citecolor=cvprblue]{hyperref}

\def\paperID{*****} 
\def\confName{CVPR}
\def\confYear{2025}

\title{NTIRE 2025 Challenge on Real-World Face Restoration: Methods and Results}

\author{Zheng Chen\thanks{Zheng Chen, Jingkai Wang, Kai Liu, Jue Gong, Lei Sun, Zongwei Wu, Radu Timofte, and Yulun Zhang are the challenge organizers, while the other authors participated in the challenge. Section~B in the supplementary materials contains the authors' teams and affiliations. NTIRE 2025 webpage: \url{https://cvlai.net/ntire/2025}. Code: \url{https://github.com/zhengchen1999/NTIRE2025_RealWorld_Face_Restoration}. } \and Jingkai Wang\footnotemark[1] \and Kai Liu\footnotemark[1] \and Jue Gong\footnotemark[1] \and Lei Sun\footnotemark[1] \and Zongwei Wu\footnotemark[1] \and Radu Timofte\footnotemark[1] \and Yulun Zhang\footnotemark[1] \thanks{Corresponding author: Yulun Zhang. \href{mailto:yulun100@gmail.com}{yulun100@gmail.com}} \and 
Jianxing Zhang \and Jinlong Wu \and Jun Wang \and Zheng Xie \and Hakjae Jeon \and Suejin Han \and Hyung-Ju Chun \and Hyunhee Park \and 
Zhicun Yin \and Junjie Chen \and Ming Liu \and Xiaoming Li \and Chao Zhou \and Wangmeng Zuo \and
Weixia Zhang \and Dingquan Li \and Kede Ma \and 
Yun Zhang \and Zhuofan Zheng \and Yuyue Liu \and Shizhen Tang \and Zihao Zhang \and Yi Ning \and Hao Jiang \and 
Wenjie An \and Kangmeng Yu \and Chenyang Wang \and Kui Jiang \and Xianming Liu \and Junjun Jiang \and 
Yingfu Zhang \and Gang He \and Siqi Wang \and Kepeng Xu \and Zhenyang Liu \and 
Changxin Zhou \and Shanlan Shen \and Yubo Duan \and 
Yiang Chen \and Jin Guo \and Mengru Yang \and 
Jen-Wei Lee \and Chia-Ming Lee \and Chih-Chung Hsu \and 
Hu Peng \and Chunming He}

\begin{document}

\maketitle

\begin{abstract}
This paper provides a review of the NTIRE 2025 challenge on real-world face restoration, highlighting the proposed solutions and the resulting outcomes. The challenge focuses on generating natural, realistic outputs while maintaining identity consistency. Its goal is to advance state-of-the-art solutions for perceptual quality and realism, without imposing constraints on computational resources or training data. The track of the challenge evaluates performance using a weighted image quality assessment (IQA) score and employs the AdaFace model as an identity checker. The competition attracted 141 registrants, with 13 teams submitting valid models, and ultimately, 10 teams achieved a valid score in the final ranking. This collaborative effort advances the performance of real-world face restoration while offering an in-depth overview of the latest trends in the field.
\end{abstract}

\vspace{-4mm}
\section{Introduction}
\vspace{-1mm}
Face restoration aims to recover high-quality (HQ) face images from low-quality (LQ) inputs that are degraded by blur, noise, compression, and other distortions. The significant loss of information makes the problem highly ill-posed. With the continuous advancement of portrait photography technology, people have higher expectations for the detail and fidelity of face images. Therefore, there is an urgent need to ensure that the restored image appears natural and realistic. Recent developments in deep learning have greatly improved face restoration. Modern approaches using CNNs, Transformers~\cite{zhou2022codeformer,wang2023restoreformer++,xie2024pltrans,tsai2024daefr}, GANs~\cite{ChenPSFRGAN,wang2021gfpgan,Yang2021GPEN,chan2021glean} and Diffusion Models~\cite{wang2023dr2,miao2024waveface,yang2023pgdiff,chen2023BFRffusion,qiu2023diffbfr,Suin2024CLRFace,wu2024osediff,lin2024diffbir,yue2024difface,tao2025overcoming,wang2025osdface}, have shown superior performance.

One of the main challenges faced by researchers is the effective capture of face priors. While statistical priors have been commonly employed in traditional image processing approaches, the rise of neural networks has led to the exploration of more learning-based methods. Geometric-prior methods~\cite{yu2018super, chen2018fsrnet, kim2019progressive, shen2018deep} are particularly useful for providing explicit facial structure information. However, for faces with less severe degradation, people prefer the output to be as realistic as possible, even with skin textures that can only be captured by high-performance cameras. Therefore, in addition to semantic guidance, texture information for restoring facial details is also important.

Many recent studies~\cite{gu2022vqfr,zhou2022codeformer,wang2023restoreformer++,xie2024pltrans,tsai2024daefr} have explored the use of Transformer models to integrate face priors. CodeFormer~\cite{zhou2022codeformer} and DAEFR~\cite{tsai2024daefr}, as two notable examples, utilize codebooks trained on HQ data as face priors. These models perform well at preserving facial information, yet are also limited in handling face and background transitions in those severely degraded images.

As for those severely degraded images, the model's generative capability becomes crucial. Many GAN-based methods~\cite{ChenPSFRGAN,wang2021gfpgan,Yang2021GPEN,chan2021glean} can effectively restore facial details. It's worth mentioning that GFPGAN~\cite{wang2021gfpgan} not only builds an effective model, but also provides many benchmark datasets for the computer vision community. In recent years, diffusion-based methods have emerged~\cite{wang2023dr2,miao2024waveface,yang2023pgdiff,chen2023BFRffusion,qiu2023diffbfr,Suin2024CLRFace,wu2024osediff,lin2024diffbir,yue2024difface,tao2025overcoming,wang2025osdface}, and with the powerful generative capabilities of diffusion models, it is now possible to restore high-quality faces from highly degraded images. DR2~\cite{wang2023dr2} contributes by transforming input images into noisy states and iteratively denoising them to capture essential semantic information. DiffBIR~\cite{lin2024diffbir}, leveraging the generative power of pre-trained LDM as a prior, greatly improves facial detail restoration. OSDFace~\cite{wang2025osdface} reduces the multi-step sampling in diffusion to a single step, achieving faster inference speeds while maintaining good image restoration quality. Additionally, super-resolution models like SUPIR~\cite{yu2024scaling} and StableSR~\cite{wang2024exploiting} have also been widely used in this competition, reflecting the strong capabilities of diffusion models in real-world face restoration.

Collaborating with the 2025 New Trends in Image Restoration and Enhancement (NTIRE 2025) workshop, we developed a challenge focused on real-world face restoration. This challenge aims to restore high-quality (HQ) face images from degraded low-quality (LQ) inputs, striving for more detailed textures and realistic face images while maintaining consistent ID. The goal is to design a network or solution that achieves high-quality restoration with the best perceptual performance, and to determine the prevailing trends in face restoration network design.


This challenge is one of the NTIRE 2025\footnote{\url{https://www.cvlai.net/ntire/2025/}} Workshop associated challenges on: ambient lighting normalization~\cite{ntire2025ambient}, reflection removal in the wild~\cite{ntire2025reflection}, shadow removal~\cite{ntire2025shadow}, event-based image deblurring~\cite{ntire2025event}, image denoising~\cite{ntire2025denoising}, XGC quality assessment~\cite{ntire2025xgc}, UGC video enhancement~\cite{ntire2025ugc}, night photography rendering~\cite{ntire2025night}, image super-resolution (x4)~\cite{ntire2025srx4}, real-world face restoration~\cite{ntire2025face}, efficient super-resolution~\cite{ntire2025esr}, HR depth estimation~\cite{ntire2025hrdepth}, efficient burst HDR and restoration~\cite{ntire2025ebhdr}, cross-domain few-shot object detection~\cite{ntire2025cross}, short-form UGC video quality assessment and enhancement~\cite{ntire2025shortugc,ntire2025shortugc_data}, text to image generation model quality assessment~\cite{ntire2025text}, day and night raindrop removal for dual-focused images~\cite{ntire2025day}, video quality assessment for video conferencing~\cite{ntire2025vqe}, low light image enhancement~\cite{ntire2025lowlight}, light field super-resolution~\cite{ntire2025lightfield}, restore any image model (RAIM) in the wild~\cite{ntire2025raim}, raw restoration and super-resolution~\cite{ntire2025raw} and raw reconstruction from RGB on smartphones~\cite{ntire2025rawrgb}.

\vspace{-1mm}
\section{NTIRE 2025 Real-world Face Restoration}
\vspace{-0.5mm}
This challenge focuses on restoring real-world degraded face images. The task is to recover high-quality face images with rich high-frequency details from low-quality inputs. At the same time, the output should preserve facial identity to a reasonable degree. There are no restrictions on computational resources such as model size or FLOPs. The main goal is to achieve the best possible image quality and identity consistency.

\subsection{Dataset}
We recommend using the FFHQ~\cite{karras2019ffhq} dataset for training, which contains 70,000 high-quality face images. Participants are also allowed to use additional datasets during the training phase. Different image sets are used for the development and testing stages. All test images are sampled from the following five datasets: 50 images from CelebChild-Test~\cite{wang2021gfpgan} and 100 images each from the LFW-Test~\cite{wang2021gfpgan}, WIDER-Test~\cite{zhou2022codeformer}, CelebA~\cite{karras2018celeba} and WebPhoto-Test~\cite{wang2021gfpgan}.

\noindent{\textbf{FFHQ.}} The FFHQ dataset contains 70,000 high-quality face images with diverse attributes and demographics. Due to its consistency and resolution, FFHQ is widely used as a training set for face restoration and generation tasks.

\noindent{\textbf{LFW-Test.}} The LFW-Test is derived from the Labeled Faces in the Wild (LFW) dataset~\cite{huang2008lfw}, consisting of 1,711 low-quality in-the-wild faces. It is formed by selecting the first image of each identity from the validation partition.

\noindent{\textbf{WIDER-Test.}} The WIDER-Test set contains 970 low-quality real-world images, sampled from the WIDER FACE dataset. It includes faces under challenging conditions such as extreme pose, occlusion, and illumination.

\noindent{\textbf{CelebChild-Test.}} The CelebChild-Test set contains 180 childhood face images of celebrities collected from the internet. CelebChild-Test features many black-and-white or low-quality images that reflect severe degradation.

\noindent{\textbf{WebPhoto-Test.}} The WebPhoto-Test set is built from 188 real-world photos crawled from the Internet, with 407 faces detected. The WebPhoto dataset features complex degradations such as aging, detail loss, and color fading.

\noindent{\textbf{CelebA.}} The CelebA dataset in the challenge is sampled from the validation set of the CelebFaces Attributes (CelebA) dataset~\cite{karras2018celeba}, which contains 19,867 images at a resolution of 178$\times$218. All images are center-cropped to 178$\times$178 and then upsampled to 512$\times$512.

\subsection{Competition} \label{sec:evaluation}
Participants are ranked based on the visual quality of their restored face images, while ensuring identity consistency with the corresponding low-quality input faces from the test set. Submissions are required to maintain identity similarity above a threshold, allowing no more than 10 low-similarity cases, and aim to maximize perceptual quality scores.

\subsubsection{Challenge Phases}
\noindent\textbf{Development and Validation Phase:} Participants are provided with 70,000 high-quality images from the FFHQ dataset, along with 450 low-quality (LQ) images sampled from five real-world datasets. By applying simulated degradations, participants can construct training pairs for supervised restoration. Additional datasets are also permitted for training. During this phase, restored high-quality images can be submitted to the Codalab evaluation server to obtain perceptual quality scores: CLIPIQA~\cite{wang2022clipiqa} and MUSIQ~\cite{ke2021musiq}.

\vspace{1mm}

\noindent\textbf{Testing Phase:} During the final testing phase, participants are given 450 new LQ test images, distinct from those used in development. Additionally, participants are required to upload their restored images to the Codalab server and email their code along with a factsheet to the organizers. The organizers will verify the submitted code and publish the final results after the challenge concludes.

\subsubsection{Evaluation Procedure}
\noindent\textbf{Step 1: Identity Similarity Measurement.} We use a pre-trained AdaFace~\cite{kim2022adaface} model to extract identity embeddings from both the input low-quality (LQ) images and the restored high-quality (HQ) images, and compute the cosine similarity between them. Due to varying degrees of degradation across datasets, different thresholds are applied for different sources. For the Wider-Test and WebPhoto-Test datasets, the threshold is set to 0.3. For the LFW-Test and CelebChild-Test datasets, the threshold is set to 0.6. For the CelebA dataset, the threshold is set to 0.5.

\vspace{0.5mm}

\noindent\textbf{Step 2: Image Quality Assessment.} We evaluate the restored HQ images using several no-reference image quality assessment (IQA) metrics: NIQE~\cite{zhang2015niqe}, CLIPIQA~\cite{wang2022clipiqa}, MANIQA~\cite{yang2022maniqa}, MUSIQ~\cite{ke2021musiq}, and Q-Align~\cite{wu2024qalign}, as well as the FID score with respect to the FFHQ dataset as a reference. To ensure fairness and reproducibility, the final ranking is mainly based on the results produced by the submitted code. These results are used for reproduction and verification. The Codalab submission serves as auxiliary confirmation, and minor differences in scores are acceptable. The evaluation scripts are available at \url{https://github.com/zhengchen1999/NTIRE2025_RealWorld_Face_Restoration}, which also includes the source code and pre-trained models of participating methods. 
The teams are ultimately ranked based on the overall perceptual score, which is computed by
\vspace{-2mm}\begin{equation*}
\begin{aligned}
   \text{Score} &= \text{CLIPIQA} + \text{MANIQA} + \frac{\text{MUSIQ}}{100} + \frac{\text{QALIGN}}{5}\\
   & + \max\left(0, \frac{10 - \text{NIQE}}{10}\right) + \max\left(0, \frac{100-\text{FID}}{100}\right). 
\end{aligned}
\end{equation*}

\begin{table*}[htbp]
\centering
\small
\begin{adjustbox}{width=\linewidth}
\begin{tabular}{cc|c|cccccc|ccc|c}
\toprule
\multirow{2}{*}{\shortstack{Team\\No.}} & \multirow{2}{*}{\shortstack{Team\\Name}} & \multirow{2}{*}{Rank} & \multirow{2}{*}{NIQE} & \multirow{2}{*}{CLIPIQA} & \multirow{2}{*}{ManIQA} & \multirow{2}{*}{MUSIQ} & \multirow{2}{*}{Q-Align} & \multirow{2}{*}{FID} & \multirow{2}{*}{\shortstack{Adaface\\Score}} & \multirow{2}{*}{\shortstack{Failed\\images}} & \multirow{2}{*}{\shortstack{ID\\Validation}} & \multirow{2}{*}{\shortstack{Total\\Score}} \\
& & & & & & & & & & & &  \\
\midrule
18 & AllForFace  & 1   & 3.9810 & 0.9517 & 0.7035 & 77.5865 & 4.4236 & 55.3807 & 0.8224 & 1   & \checkmark & 4.3638 \\
7  & IIL         & 2   & 3.6979 & 0.9337 & 0.6826 & 78.4876 & 4.2337 & 52.2038 & 0.8503 & 1   & \checkmark & 4.3561 \\
9  & PISA-MAP    & 3   & 4.5488 & 0.9480 & 0.6603 & 78.8849 & 4.5209 & 57.3616 & 0.7904 & 2   & \checkmark & 4.2728 \\
10 & MiPortrait  & 4   & 5.8223 & 0.9729 & 0.7725 & 78.2018 & 4.6092 & 64.3907 & 0.7452 & 5   & \checkmark & 4.2231 \\
2  & AIIA        & 5   & 4.2946 & 0.9046 & 0.6559 & 77.0162 & 4.2834 & 60.0422 & 0.7497 & 2   & \checkmark & 4.1575 \\
14 & UpHorse     & 6   & 5.7813 & 0.8834 & 0.7444 & 76.0564 & 4.5482 & 61.0268 & 0.7303 & 2   & \checkmark & 4.1095 \\
11 & CX          & 7   & 5.3603 & 0.7780 & 0.6336 & 75.2278 & 4.3585 & 53.4188 & 0.8388 & 3   & \checkmark & 3.9654 \\
16 & AIIALab     & 8   & 5.0737 & 0.7554 & 0.6231 & 75.7016 & 4.3272 & 56.5300 & 0.6541 & 10  & \checkmark & 3.9283 \\
8  & ACVLab      & 9   & 6.0577 & 0.8077 & 0.6951 & 76.8980 & 4.3840 & 62.2699 & 0.7539 & 1   & \checkmark & 3.9201 \\
4  & HH          & 10  & 6.3267 & 0.8000 & 0.6591 & 75.6512 & 4.3302 & 60.1384 & 0.7934 & 3   & \checkmark & 3.8475 \\
\midrule
3  & Fustar-fsr  & N/A & 4.2674 & 0.7933 & 0.6605 & 77.6443 & 4.3848 & 59.9098 & 0.4485 & 229 & $\times$   & 4.0814 \\
6  & Night Watch & N/A & 6.5009 & 0.8593 & 0.7786 & 77.1615 & 4.5075 & 76.5814 & 0.5348 & 130 & $\times$   & 3.8952 \\
17 & IPCV        & N/A & 5.4568 & 0.6781 & 0.5298 & 71.0464 & 4.1491 & 55.7525 & 0.6380 & 17  & $\times$   & 3.6450 \\
\midrule
\end{tabular}
\end{adjustbox}
\caption{Results of NTIRE 2025 Real-world Face Restoration Challenge. The testing was conducted on the test dataset, consisting of 450 images from CelebChild-Test, LFW-Test, WIDER-Test, CelebA, and WebPhoto-Test. Participants were required to pass the AdaFace ID Test first to qualify for ranking. The final results were calculated based on the weighted score of no-reference IQA metrics for the ranking.}
\label{tab:main_results}
\end{table*}

\vspace{-4mm}
\section{Challenge Results}
Table~\ref{tab:main_results} presents the final rankings and results of the teams. A comprehensive description of the evaluation process is outlined in Sec.~\ref{sec:evaluation}. The top 5 teams, along with their method details, are provided in Sec.~\ref{sec:teams}, while the remaining teams are listed in Sec.~A of the supplementary materials. In addition, team member information can be found in Sec.~B of the supplementary materials. The AllForFace team achieved the top ranking in this challenge. Furthermore, three teams did not pass the AdaFace ID test, and as a result, their scores are not deemed valid.

\subsection{Architectures and main ideas}
Throughout the challenge, various novel approaches were proposed to enhance the performance of face restoration. In the following, we highlight some key ideas observed in the submitted solutions.

\begin{enumerate}

    \item \textbf{Diffusion-based generative priors are widely utilized.} 
    Diffusion models have become the primary choice for enhancing high-frequency details and realism in face restoration. Many participants leveraged a two-stage framework, first removing degradations via CNN-based or Transformer-based restorers and subsequently refining results using diffusion-based generative modules such as DiffBIR~\cite{lin2024diffbir}. For example, the IIL team further explored a highly efficient one-step diffusion model (SDFace) distilled from SDXL-Turbo, achieving fast inference with strong perceptual performance through end-to-end diffusion generation.

    \vspace{1mm}

    \item \textbf{Incorporating Transformer-based priors improves identity consistency.} 
    Preserving facial identity was a core objective, prompting participants to integrate Transformer-based modules like CodeFormer~\cite{zhou2022codeformer} or Vision Transformers. These methods provided robust structural guidance, improving identity preservation even in severely degraded scenarios. For example, the ACVLab team combined CodeFormer and DiffBIR in a two-stage pipeline, leveraging the Transformer prior in the first stage to enhance structural fidelity under diverse degradation conditions.

    \vspace{1mm}

    \item \textbf{Fusing diverse generative priors for progressive face restoration.}
    Some participants explored integrating multiple generative priors—such as GANs, diffusion models, VAEs, and Transformer-based modules—to tackle different aspects of face restoration. This year's winning team, SRC-B, adopted a three-stage pipeline that sequentially applied StyleGAN for identity fidelity, a diffusion model for texture realism, and a VAE guided by vision foundation models for naturalness. This progressive fusion allowed targeted optimization across stages, achieving high perceptual quality and structural consistency.

    \vspace{1mm}

    \item \textbf{Dynamic adaptive weighting of multiple restoration modules.} 
    Some innovative approaches involved adaptive architectures employing conditional gating or Mixture-of-Experts (MoE) frameworks. These methods dynamically weighted different restoration modules based on image degradation severity, ensuring optimal balance between perceptual quality and identity fidelity. For example, the AIIALab team proposed a dual-model MoE framework that adaptively fuses the outputs of DiffBIR and CodeFormer using a lightweight gating network, enabling flexible control over structural fidelity and texture realism.

    \item \textbf{Iterative latent-space optimization provides image-specific refinement.} 
    Advanced methods applied iterative optimization techniques in the diffusion latent space, guided by no-reference IQA metrics such as CLIPIQA~\cite{wang2022clipiqa} and MUSIQ~\cite{ke2021musiq}. This approach provided effective per-image fine-tuning, significantly boosting subjective quality. For example, the PISA-MAP team employed MAP estimation in the diffusion latent space, iteratively optimizing each image under the guidance of multiple NR-IQA metrics, which effectively enhanced perceptual quality in a content-aware manner.
\end{enumerate}

\vspace{-1.mm}
\subsection{Participants}
\vspace{-1.mm}
This year, the real-world face restoration challenge saw 141 registered participants, with 13 teams submitting their valid models, and finally, 10 teams had a valid score in the final ranking. These entries set a new standard for the state-of-the-art in face restoration.

\vspace{-1.mm}
\subsection{Fairness}
\vspace{-1.mm}
A set of rules is instituted to maintain the fairness of the competition. \textbf{(1)} The use of FFHQ dataset for training, with no overlapping images from the five test datasets: LFW-Test, WIDER-Test, CelebChild-Test, CelebA, and WebPhoto-Test, is required. \textbf{(2)} Training with additional datasets, e.g., FFHQR, is permitted. \textbf{(3)} The application of no-reference IQA and simulated degradation pipelines during training and testing is considered a fair practice.

\vspace{-1.mm}
\subsection{Conclusions}
\vspace{-1.mm}
The insights gained from analyzing the results of the face restoration challenge are summarized as follows:
\begin{enumerate}
    \item This competition has driven advancements in the field of face restoration, with the proposed techniques significantly improving performance in practical applications.
    \item Diffusion-based approaches are capable of generating highly realistic face images, while Transformer-based methods have demonstrated excellent performance in maintaining facial features and achieving high fidelity.
    \item By combining multiple models and feature extraction methods and performing joint optimization, the final results yield natural and realistic face images.
\end{enumerate}

\vspace{-1.mm}
\section{Challenge Methods and Teams}
\label{sec:teams}

\subsection{AllForFace}
\noindent\textbf{Description.} In real-world scenarios, face images may suffer from various types of degradation, such as noise,
blur, down-sampling, JPEG compression artifacts, and etc. Blind face restoration (BFR) aims to
restore high-quality face images from low-quality ones that suffer from unknown degradation. The
BFR approaches are mainly focused on exploring better face priors, including geometric priors~\cite{chen2018fsrnet, ChenPSFRGAN}, reference priors~\cite{zhou2022codeformer}, and generative priors~\cite{wang2021gfpgan}. Diffusion prior~\cite{lin2024diffbir, yu2024scaling}, which is more explored in recent years, belongs to a broader stream of generative priors.

\begin{figure}[t]
    \centering
    \includegraphics[width=0.7\linewidth]{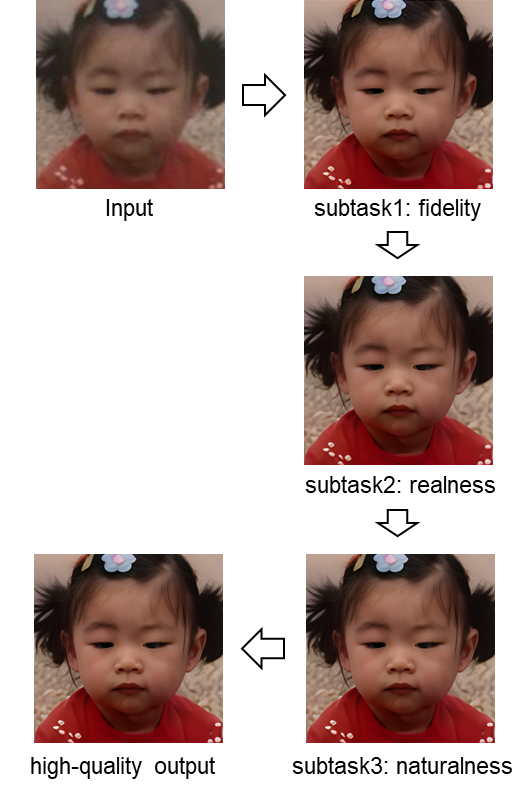}
    \caption{\textbf{Team AllForFace.} Basic idea. }
    \label{fig:team18fig1}
\end{figure}

Currently, solutions for blind face recovery primarily address the problem from two angles: one focuses on "how degradation occurs," simulating more realistic degradation patterns by concentrating on degrading factors. The other considers exploitable prior information, i.e., which information can be utilized to supplement low-quality images. However, their proposed solution addresses the issue from a user-centric perspective. Specifically, they decompose the blind face recovery problem into three sub-problems corresponding to users' three fundamental requirements: fidelity (faithful reconstruction), realness (authentic representation), and naturalness (natural appearance). For each sub-problem, they design distinct solutions and integrate them into a serial pipeline to achieve high-quality reconstructed images. Fig.~\ref{fig:team18fig1} demonstrates their divide-and-conquer approach.

\begin{figure}[t]
    \centering
    \includegraphics[width=0.8\linewidth]{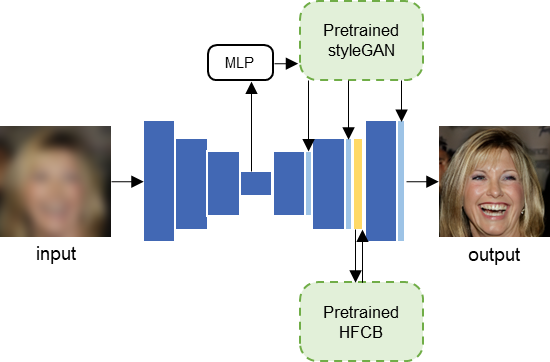}
    \caption{\textbf{Team AllForFace.} Fidelity mode structure. }
    \label{fig:team18fig2}
\end{figure}

\noindent\textbf{Implementation Details.} The first subtask aims to achieve faithfulness, which is the most fundamental requirement in face restoration, i.e., maintaining a high degree of consistency between the restored identity and the true identity. To address this challenge, the pre-trained face generation model, StyleGANv2, was utilized to provide enhanced generative priors for the face restoration process. While diffusion-based priors have become a more widely used category of generative priors in recent years, their generation capabilities become difficult to control under weaker conditioning (e.g., severely degraded faces). As illustrated in Fig.~\ref{fig:team18fig2}, their model design incorporates high-quality face priors extracted from StyleGANv2 based on facial pre-training and integrates them into the decoding process, where they are jointly decoded with the original image features. Additionally, additional face component priors were extracted from the Pretrained High-Quality Face Component Basis (HFCB) model to provide finer component-based priors for the current face restoration process. The fidelity model ensures that, even under severe facial degradation, it maintains the best consistency with the input in terms of fine texture, color, and shape.

In addition to achieving consistent ID information, it is crucial to further enhance the details of the texture to render them more realistic. For instance, when facial degradation is extremely severe, the restored hair strands using GAN-based methods often fail to conform to realistic hair strand patterns. To address the second subtask, diffusion models were prioritized for consideration. Since the image has already achieved a high degree of ID consistency following the first subtask, this provides strong control information for diffusion models. Leveraging the prior knowledge embedded in the diffusion process enables simultaneous enhancement of texture details while maintaining ID consistency. Specifically, they improved upon existing work DiffBIR~\cite{lin2024diffbir} by explicitly removing its first stage and retaining only the diffusion stage to facilitate more realistic texture recovery for facial restoration.

\begin{figure}[t]
    \centering
    \includegraphics[width=1.0\linewidth]{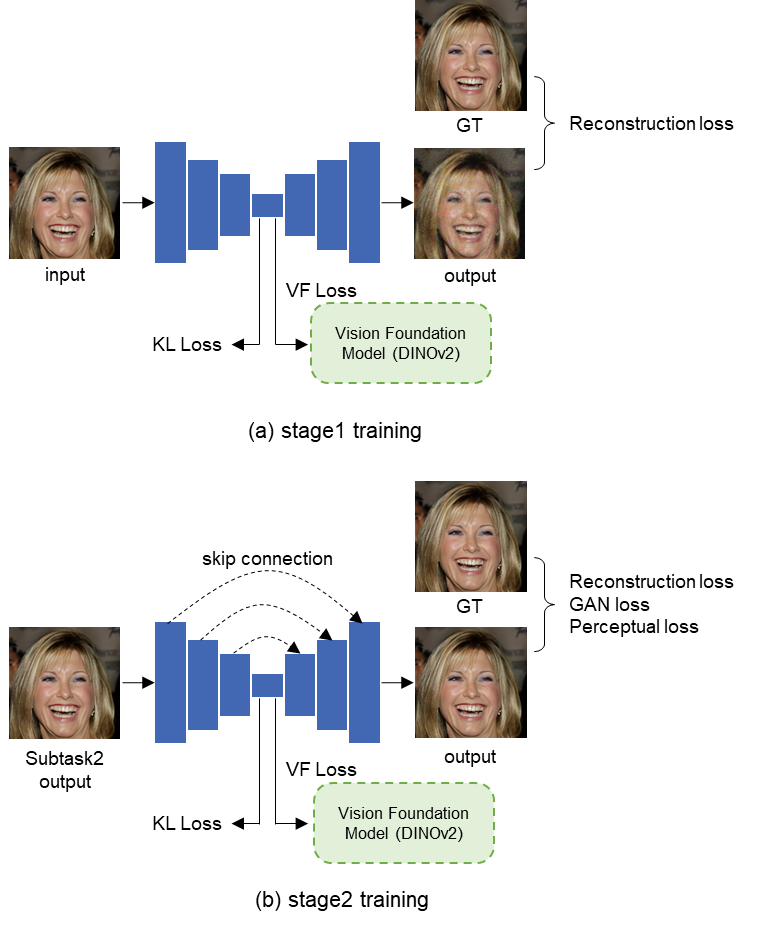}
    \caption{\textbf{Team AllForFace.} Naturalness model. }
    \label{fig:team18fig3}
    \vspace{2.mm}
\end{figure}

After completing the two subtasks, they obtained facial images with improved texture details. However, upon observation, it became evident that some of the restored faces did not align with the statistical distribution characteristics of natural images. For example, certain restored faces exhibit overly delicate hair strands and excessively smooth skin, despite the improved clarity and detailed information in these images. However, overall, they lack a sense of naturalness in their appearance. To address this, the third subtask was designed to transform the restoration process into an image reconstruction task, aiming to convert texture-rich but unnatural images into texture-rich and natural images.

Specifically, this subtask consists of two stages. In the first stage, instead of using traditional VAE (Variational Autoencoder) models, they incorporated vision foundation models, such as Dinov2~\cite{oquab2023dinov2}, to guide the optimization of the VAE latent space. This approach significantly enhances the naturalness of face images while preserving their original reconstruction capabilities (see Fig.~\ref{fig:team18fig3}). The integration of vision foundation models enables more sophisticated control over the reconstruction process, ensuring that the resulting images not only retain fine texture details but also conform to the statistical properties of natural images. In the second stage, they maintain the core network architecture while introducing skip connections between the encoder and decoder. This modification added additional reconstruction loss, perceptual loss, and GAN (Generative Adversarial Network) loss terms to the training objective. The purpose of these additions was to preserve the texture details while encouraging the generation of more natural-looking facial images. It is important to note that the input to this stage was the output of the second subtask, ensuring that the final results achieved improved naturalness without sacrificing texture details.

\begin{figure}[t]
	\centering
        \footnotesize
	\includegraphics[width=1.0\linewidth]{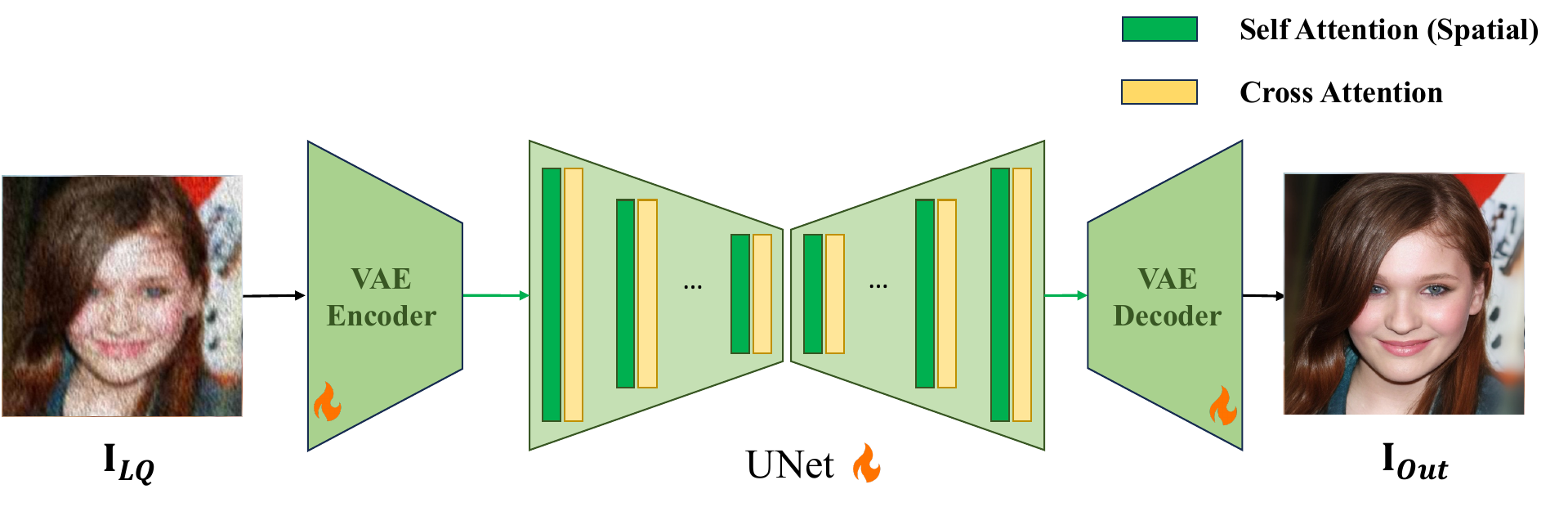}
    \caption{\textbf{Team IIL.} }
    \label{fig:network}
\end{figure}

\begin{figure}[t]
    \centering
    \includegraphics[width=\columnwidth]{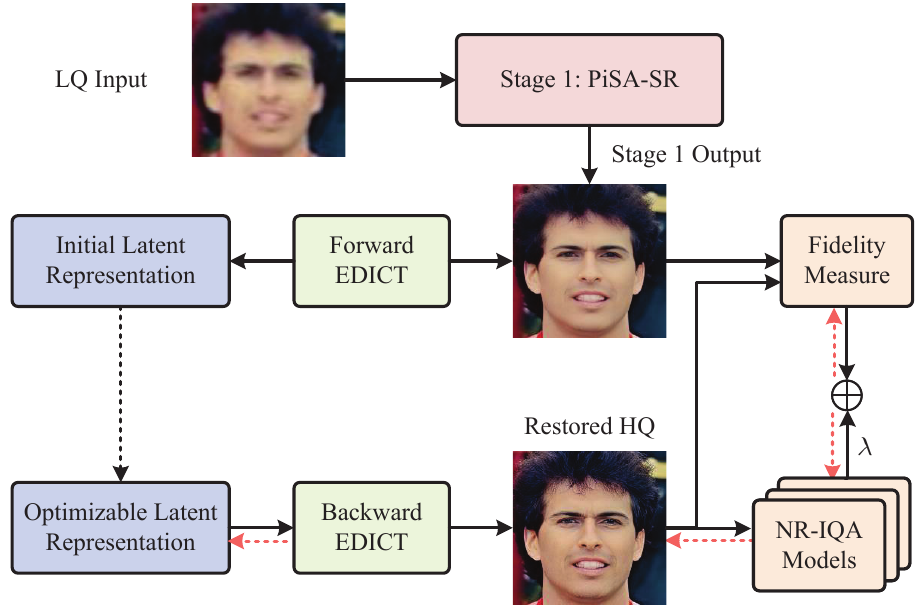}
    \caption{\textbf{Team PISA-MAP.}}
    \label{fig:enter-label}
\end{figure}

\subsection{IIL}
\noindent\textbf{Description.} Their framework extends the pre-trained Stable Diffusion XL model~\cite{SDXL}. This foundation model generates photorealistic facial variations based on its strong prior. 
Attracted by the enhanced restoration performance and fast processing speed of the one-step diffusion framework~\cite{wang2025osdface, InstantRestore}, they adopt the diffusion model as a single-step face image restoration model.
To turn the base model into a one-step diffusion framework, they utilize the distilled SDXL Turbo model~\cite{SDXL-Turbo} and design a set of LoRA adapters for both VAE and UNet to build a one-step diffusion model SDFace for face restoration, shown in Fig.~\ref{fig:network}.

\begin{figure*}[t]
    \centering
    \includegraphics[width=1\linewidth]{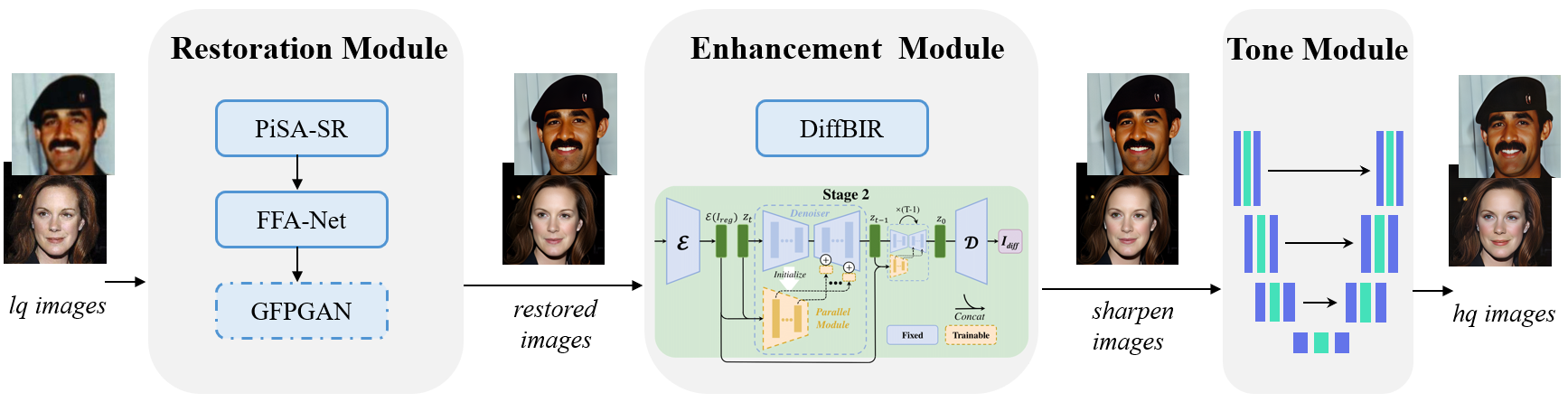}
    \caption{\textbf{Team MiPortrait.}}
    \label{Fig.1}
\end{figure*}

\noindent\textbf{Implementation Details.} Based on the one-step diffusion framework, they directly use image-level loss functions for the supervision. Their loss function contains fidelity loss ($\ell_1$), perception loss~\cite{wang2021real}, ID loss~\cite{deng2019arcface} and GAN loss~\cite{chen2022real}.
In specific, the fidelity loss is
\begin{equation}
\mathcal{L}_1=\|\mathbf{I}_\mathit{Out}-\mathbf{I}\|_1,
\end{equation}
where $\mathbf{I}$ and $\mathbf{I}_\mathit{Out}$ denotes the ground truth face image and restored output, and the perception loss is defined by
\begin{equation}
    \mathcal{L}_\mathrm{perc} = \sum_{i}^{n}\|\phi_{i}(\mathbf{I}_\mathit{Out})-\phi_{i}(\mathbf{I})\|_2,
\end{equation}
where $\phi$ is the pre-trained VGG~\cite{simonyan2015vgg} feature extractor and $i$ denotes the i th layer of VGG.
The ID loss is defined by,
\begin{equation}
    \mathcal{L}_\mathrm{id} = 1 -  \langle R(\mathbf{I}_\mathit{Out})-R(\mathbf{I}) \rangle,
\end{equation}
where $R$ is the face recognition network ArcFace~\cite{deng2019arcface}. 
%
The adversarial loss follows the setting of Real-ESRGAN~\cite{chen2022real}, which is defined by,
\begin{equation}
    \mathcal{L}_\mathrm{adv} = -\mathbb{E}[\log(\mathit{D}(\mathbf{I}_\mathit{Out}))],
\end{equation}
where the discriminator $\mathit{D}$ is iteratively trained along with the restoration network,
\begin{equation}
    \mathcal{L}_\mathit{D} = -\mathbb{E}[\log(\mathit{D}(\mathbf{I}))-\log(1-\mathit{D}(\mathbf{I}_\mathit{Out}))].
    \label{eqn:RefSTAR_loss_D}
\end{equation}

Overall, the formula for the final learning objective is:
\begin{equation}
    \mathcal{L} = \lambda_1\mathcal{L}_1 + \lambda_2\mathcal{L}_\mathrm{perc} +
    \lambda_3\mathcal{L}_\mathrm{id} + \lambda_4\mathcal{L}_\mathrm{adv},
\end{equation}
where the hyperparameters $\lambda_1$, $\lambda_2$, $\lambda_3$, and $\lambda_4$ are empirically set to 1, 1, 1, and 0.5.

The training process of their SDFace uses the AdamW optimizer with $\beta_1$ = 0.5 and $\beta_2$ = 0.999, with a learning rate of $1\times10^{-4}$.
All experiments were conducted on a server equipped with two NVIDIA A6000 GPUs. 

The model uses pretrained SDXL-Turbo~\cite{SDXL-Turbo}, and has 2703.64 M parameters, 2771.52 GFLOPS, and 0.21s per image on one NVIDIA A6000 GPU for inference.

\subsection{PISA-MAP}
\noindent\textbf{Description.} Their method follows a two-stage restoration design philosophy, showing in Fig.~\ref{fig:enter-label}. The first stage focuses on degradation removal using a feedforward image enhancer, while the second stage applies maximum a posteriori (MAP) estimation in Diffusion Latents~\cite{zhang2024no} for post-enhancement, iteratively refining the outputs from the first stage. For the first stage, they use PiSA-SR~\cite{sun2024pixel} as the image enhancer. In their preliminary experiments, they also tested the larger SUPIR model~\cite{yu2024scaling}. However, they found that SUPIR tends to generate unrealistic details (i.e., hallucinations), which is undesirable for real-world face restoration, where maintaining identity consistency is as important as achieving high perceptual quality. In the second stage, they follow the approach in~\cite{zhang2024no}, with one modification: they employ three NR-IQA models to jointly guide the end-to-end optimization of the latents inverted from input images using EDICT~\cite{wallace2023edict}. Specifically, they use a multi-scale variant of LIQE~\cite{zhang2023blind} as proposed in~\cite{zhang2024no}, alongside MUSIQ~\cite{ke2021musiq} and CLIP-IQA~\cite{wang2022clipiqa}.

\noindent\textbf{Implementation Details.} All the modules in their method, including the PiSA-SR~\cite{sun2024pisasr} model used in the first stage, and the NR-IQA models along with the backbone diffusion model used in the second stage, are trained according to their original schemes~\cite{wu2024seesr}. They do \textbf{not} conduct any additional training of model parameters, but perform instance-wise iterative MAP estimation of diffusion latents in the second stage. The data they used are LSDIR~\cite{li2023lsdir} and the first 10K images from FFHQ~\cite{karras2019ffhq}. Paired training data are generated using the degradation pipeline from RealESRGAN~\cite{wang2021real}. MS-LIQE is jointly trained on six IQA datasets: LIVE~\cite{sheikh2006statistical}, CSIQ~\cite{larson2010most}, KADID-10k~\cite{lin2019kadid}, BID~\cite{ciancio2011no}, LIVE Challenge~\cite{ghadiyaram2016massive}, KonIQ-10k~\cite{hosu2020koniq}. 

As for testing, they follow the default setup of PiSA-SR~\cite{sun2024pisasr} in the first stage. In the second stage, they configure EDICT with a mixing parameter of $p = 0.93^{50/T}$ and set the number of diffusion steps to $T = 20$.  Each image is optimized for up to $\mathrm{MaxIter}= 30$ iterations, with a learning rate $\alpha = 2$ and a momentum parameter $\rho = 0.9$.

\subsection{MiPortrait}

\begin{figure*}[t]
\centering
\includegraphics[width=0.8\linewidth]{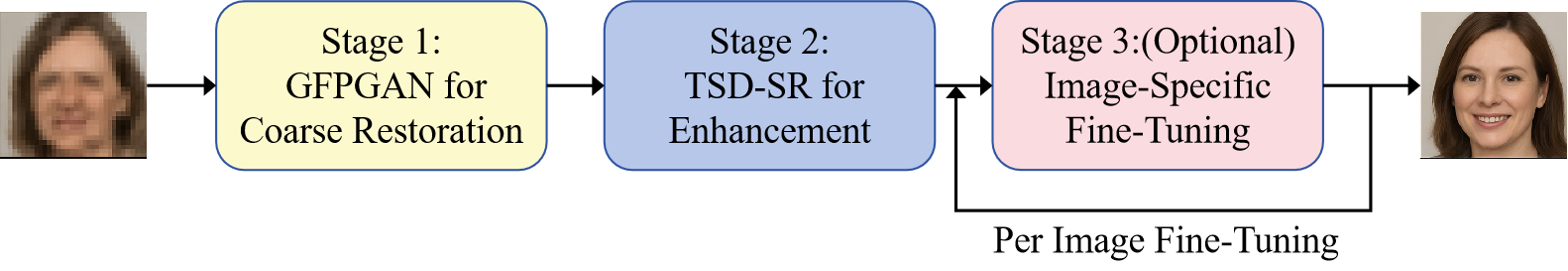}
\vspace{-0.4cm}  %
\caption{\textbf{Team AIIA.}}
\label{fig:team02-model}
\vspace{-0.4cm} 
\end{figure*}

\noindent\textbf{Description.} Today, Stable Diffusion shows unrivaled detail generation capabilities in the field of image generation. They utilize DiffBIR~\cite{lin2024diffbir} as their algorithm backbone, which adopts a two-stage framework. In the first stage, a restoration module, such as SwinIR~\cite{liang2021swinir}, is used to remove most of the degradations in the image, such as noise and blurring. In the second stage, it takes advantage of the powerful generative ability of Stable Diffusion.

\noindent\textbf{Implementation Details.} Due to the severe degradation of the images, they found that SwinIR in stage 1 could not provide sufficient facial details for stage 2. Therefore, SwinIR~\cite{liang2021swinir} is replaced by PiSA-SR\cite{sun2024pixel}, a fast one-step SD network with high fidelity.

Furthermore, in the Wider-test and Webphoto-test images, the faces are so severely degraded that they are barely recognizable even to the human eye. To address this challenge, they incorporate GFPGAN~\cite{wang2021gfpgan} at the end of the Restoration Module to render these faces more human-like. However, GFPGAN~\cite{wang2021gfpgan} has a tendency to reduce the consistency with the original image. Thus, they decided to apply it only to images of extremely low quality. CLIPIQA~\cite{wang2022clipiqa} and MUSIQ~\cite{ke2021musiq} are employed as thresholds to determine whether GFPGAN~\cite{wang2021gfpgan} should be activated.

All the processes described above are designed to enhance image sharpness. Meanwhile, tone and clarity are also crucial factors in Image Aesthetic Assessment. Due to rephotographing, sunlight exposure, and air oxidation, the color of many old photos has deteriorated. To address this, the dehaze network FAA-Net~\cite{qin2019ffanetfeaturefusionattention} is implemented to mitigate the whitening effect. At the end of the entire process, a skin-tone-specific UNet is utilized. This application aims to render the face more natural-looking and ensure it better aligns with their aesthetic criteria. The workflow of the proposed algorithm is illustrated in Fig. \ref{Fig.1}.

\subsection{AIIA}
\noindent\textbf{Description.} They adopt a three-stage face restoration pipeline that first applies GFPGAN for coarse restoration, then uses TSD-SR for high-quality enhancement, and finally incorporates ZSSR for fine-tuning the restoration process. This enhanced pipeline aims to improve the face image quality by sequentially leveraging the strengths of a GAN-based face prior, a diffusion-based super-resolution model, and a zero-shot super-resolution approach that performs fine-tuning on a per-image basis, shown in Fig.~\ref{fig:team02-model}.

\noindent\textbf{Implementation Details.} 

\noindent\textit{Stage 1: GFPGAN for Coarse Restoration}:
GFPGAN \cite{wang2021gfpgan} serves as the first stage of their pipeline, which performs coarse restoration of facial structures. It utilizes a generative facial prior network to recover identity, structure, and base details from a low-quality input image. This helps reconstruct the face’s essential features, even when large degradations are present.

\noindent\textit{Stage 2: TSD-SR for High-Quality Enhancement}:
TSD-SR \cite{dong2024tsd} is employed in the second stage to enhance the fine details of the face image. This model uses a diffusion-based super-resolution approach to inject realistic textures, refine facial details, and improve the overall quality of the restored image. By utilizing the powerful diffusion model, they can enhance the appearance of fine structures, such as wrinkles, skin textures, and other subtle features, which are difficult to recover using traditional methods.

\noindent\textit{Stage 3: ZSSR for Image-Specific Fine-Tuning}:
In the third stage, they apply the ZSSR (Zero-Shot Super-Resolution) approach \cite{shocher2018zero} to perform image-specific fine-tuning. ZSSR is a powerful technique that operates on each image individually, allowing us to refine the restoration further based on the unique characteristics of the image. This approach performs super-resolution without additional training on external datasets, utilizing the specific features of the input image to guide the restoration process. This step of the ZSSR method is optional. If you want to quickly generate the restored image, you can adopt the strategy of Stage 1 and Stage 2 only, which is also excellent.

\noindent\textit{Loss Function}:
To improve the perceptual quality of the restoration, they introduce the CLIP-IQA \cite{wang2022clipiqa} loss function. This perceptual loss is designed to optimize the alignment of the restored image with human subjective preferences. CLIP-IQA uses CLIP embeddings to measure perceptual similarity between the ground truth and the restored image.

The overall objective function $\mathcal{L}$ is defined as:
\begin{equation}
\mathcal{L} =  \mathcal{L}_{\text{CLIP-IQA}}(I, \hat{I}),
\end{equation}
where $I$ is the ground truth high-quality image, and $\hat{I}$ is the predicted restored image.
$\mathcal{L}_{\text{CLIP-IQA}}(I, \hat{I})$ is the CLIP-IQA perceptual loss, focusing on perceptual quality.

\noindent\textit{Pipeline Overview}:
GFPGAN~\cite{wang2021gfpgan} performs coarse restoration, recovering facial identity and removing common degradations.
TSD-SR~\cite{dong2024tsd} refines details, adding realistic textures and fine structures.
ZSSR~\cite{shocher2018zero} fine-tunes each image individually, further enhancing details specific to the input image.
CLIP-IQA \cite{wang2022clipiqa} is used as the perceptual loss function to guide the restoration and ensure high perceptual quality in the final result.

\noindent\textit{Training Strategy}:
GFPGAN~\cite{wang2021gfpgan} uses official pretrained weights, without any additional fine-tuning, leveraging the strength of its generative facial prior network.
TSD-SR~\cite{dong2024tsd} was trained on diverse datasets (FFHQ, DIV2K, Flickr2K, LSDIR) using realistic image degradations. LoRA adapters were applied for efficient training and inference, guided by reconstruction loss, knowledge distillation, and ground-truth supervision. They use the official pretrained model for inference without further fine-tuning.
ZSSR~\cite{shocher2018zero} is applied in a zero-shot manner, meaning that it performs per-image enhancement directly, without the need for training on external datasets. This allows it to adapt to the specific characteristics of the input image, providing fine-tuning and high-quality restoration tailored to each case.

\section{Methods of the Remaining Teams}
All the teams presented innovative ideas and thorough experiments for the competition. However, due to space limitations, a more in-depth discussion can be found in Sec.~A of the supplementary materials, which contains detailed descriptions of the methods and implementation details for the remaining teams that participated in the challenge. While these teams have not been discussed in the main report, their approaches are still highlighted, offering insight into their unique strategies and technical implementations.

\vspace{-2.mm}
\section*{Acknowledgements}
\vspace{-2.mm}
This work is supported by the Shanghai Municipal Science and Technology Major Project (2021SHZDZX0102) and the Fundamental Research Funds for the Central Universities.
This work was partially supported by the Humboldt Foundation. We thank the NTIRE 2025 sponsors: ByteDance, Meituan, Kuaishou, and University of Wurzburg (Computer Vision Lab).

{\small
\bibliographystyle{ieeenat_fullname}
\bibliography{main}

\begin{thebibliography}{84}
\providecommand{\natexlab}[1]{#1}
\providecommand{\url}[1]{\texttt{#1}}
\expandafter\ifx\csname urlstyle\endcsname\relax
  \providecommand{\doi}[1]{doi: #1}\else
  \providecommand{\doi}{doi: \begingroup \urlstyle{rm}\Url}\fi

\bibitem[Chan et~al.(2021)Chan, Wang, Xu, Gu, and Loy]{chan2021glean}
Kelvin~CK Chan, Xintao Wang, Xiangyu Xu, Jinwei Gu, and Chen~Change Loy.
\newblock Glean: Generative latent bank for large-factor image
  super-resolution.
\newblock In \emph{CVPR}, 2021.

\bibitem[Chen et~al.(2021)Chen, Li, Lingbo, Lin, Zhang, and Wong]{ChenPSFRGAN}
Chaofeng Chen, Xiaoming Li, Yang Lingbo, Xianhui Lin, Lei Zhang, and
  Kwan-Yee~K. Wong.
\newblock Progressive semantic-aware style transformation for blind face
  restoration.
\newblock In \emph{CVPR}, 2021.

\bibitem[Chen et~al.(2022)Chen, Shi, Qin, Li, Han, Yang, and Guo]{chen2022real}
Chaofeng Chen, Xinyu Shi, Yipeng Qin, Xiaoming Li, Xiaoguang Han, Tao Yang, and
  Shihui Guo.
\newblock Real-world blind super-resolution via feature matching with implicit
  high-resolution priors.
\newblock In \emph{ACM MM}, 2022.

\bibitem[Chen et~al.(2023)Chen, Tan, Wang, Zhang, Luo, and
  Cao]{chen2023BFRffusion}
Xiaoxu Chen, Jingfan Tan, Tao Wang, Kaihao Zhang, Wenhan Luo, and Xiaocun Cao.
\newblock Towards real-world blind face restoration with generative diffusion
  prior.
\newblock \emph{arXiv preprint arXiv:2312.15736}, 2023.

\bibitem[Chen et~al.(2018)Chen, Tai, Liu, Shen, and Yang]{chen2018fsrnet}
Yu Chen, Ying Tai, Xiaoming Liu, Chunhua Shen, and Jian Yang.
\newblock Fsrnet: End-to-end learning face super-resolution with facial priors.
\newblock In \emph{CVPR}, 2018.

\bibitem[Chen et~al.(2025{\natexlab{a}})Chen, Liu, Gong, Wang, Sun, Wu,
  Timofte, Zhang, et~al.]{ntire2025srx4}
Zheng Chen, Kai Liu, Jue Gong, Jingkai Wang, Lei Sun, Zongwei Wu, Radu Timofte,
  Yulun Zhang, et~al.
\newblock {NTIRE} 2025 challenge on image super-resolution (×4): Methods and
  results.
\newblock In \emph{Proceedings of the IEEE/CVF Conference on Computer Vision
  and Pattern Recognition (CVPR) Workshops}, 2025{\natexlab{a}}.

\bibitem[Chen et~al.(2025{\natexlab{b}})Chen, Wang, Liu, Gong, Sun, Wu,
  Timofte, Zhang, et~al.]{ntire2025face}
Zheng Chen, Jingkai Wang, Kai Liu, Jue Gong, Lei Sun, Zongwei Wu, Radu Timofte,
  Yulun Zhang, et~al.
\newblock {NTIRE} 2025 challenge on real-world face restoration: Methods and
  results.
\newblock In \emph{Proceedings of the IEEE/CVF Conference on Computer Vision
  and Pattern Recognition (CVPR) Workshops}, 2025{\natexlab{b}}.

\bibitem[Ciancio et~al.(2011)Ciancio, {Targino da Costa}, {da Silva}, Said,
  Samadani, and Obrador]{ciancio2011no}
Alexandre Ciancio, A.~L. N.~T. {Targino da Costa}, E.~A.~B. {da Silva}, Amir
  Said, Ramin Samadani, and Pere Obrador.
\newblock No-reference blur assessment of digital pictures based on
  multifeature classifiers.
\newblock \emph{IEEE TIP}, 2011.

\bibitem[Conde et~al.(2025{\natexlab{a}})Conde, Timofte, et~al.]{ntire2025raw}
Marcos Conde, Radu Timofte, et~al.
\newblock {NTIRE} 2025 challenge on raw image restoration and super-resolution.
\newblock In \emph{Proceedings of the IEEE/CVF Conference on Computer Vision
  and Pattern Recognition (CVPR) Workshops}, 2025{\natexlab{a}}.

\bibitem[Conde et~al.(2025{\natexlab{b}})Conde, Timofte,
  et~al.]{ntire2025rawrgb}
Marcos Conde, Radu Timofte, et~al.
\newblock Raw image reconstruction from {RGB} on smartphones. {NTIRE} 2025
  challenge report.
\newblock In \emph{Proceedings of the IEEE/CVF Conference on Computer Vision
  and Pattern Recognition (CVPR) Workshops}, 2025{\natexlab{b}}.

\bibitem[Deng et~al.(2019)Deng, Guo, Niannan, and Zafeiriou]{deng2019arcface}
Jiankang Deng, Jia Guo, Xue Niannan, and Stefanos Zafeiriou.
\newblock {ArcFace}: Additive angular margin loss for deep face recognition.
\newblock In \emph{CVPR}, 2019.

\bibitem[Dong et~al.(2024)Dong, Fan, Guo, Wang, Zhang, Chen, Luo, and
  Zou]{dong2024tsd}
Linwei Dong, Qingnan Fan, Yihong Guo, Zhonghao Wang, Qi Zhang, Jinwei Chen,
  Yawei Luo, and Changqing Zou.
\newblock Tsd-sr: One-step diffusion with target score distillation for
  real-world image super-resolution.
\newblock \emph{arXiv preprint arXiv:2411.18263}, 2024.

\bibitem[Ershov et~al.(2025)Ershov, Korchagin, Khalin, Panshin, Terekhin,
  Zaychenkova, Lobarev, Plokhotnyuk, Abramov, Zhdanov, Dorogova, Mamedov,
  Banic, Perevozchikov, Timofte, et~al.]{ntire2025night}
Egor Ershov, Sergey Korchagin, Alexei Khalin, Artyom Panshin, Arseniy Terekhin,
  Ekaterina Zaychenkova, Georgiy Lobarev, Vsevolod Plokhotnyuk, Denis Abramov,
  Elisey Zhdanov, Sofia Dorogova, Yasin Mamedov, Nikola Banic, Georgii
  Perevozchikov, Radu Timofte, et~al.
\newblock {NTIRE} 2025 challenge on night photography rendering.
\newblock In \emph{Proceedings of the IEEE/CVF Conference on Computer Vision
  and Pattern Recognition (CVPR) Workshops}, 2025.

\bibitem[Fu et~al.(2025)Fu, Qiu, Fu, Timofte, Sebe, Yang, Van~Gool,
  et~al.]{ntire2025cross}
Yuqian Fu, Xingyu Qiu, Bin Ren~Yanwei Fu, Radu Timofte, Nicu Sebe, Ming-Hsuan
  Yang, Luc Van~Gool, et~al.
\newblock {NTIRE} 2025 challenge on cross-domain few-shot object detection:
  Methods and results.
\newblock In \emph{Proceedings of the IEEE/CVF Conference on Computer Vision
  and Pattern Recognition (CVPR) Workshops}, 2025.

\bibitem[Ghadiyaram and Bovik(2016)]{ghadiyaram2016massive}
Deepti Ghadiyaram and Alan~C. Bovik.
\newblock Massive online crowdsourced study of subjective and objective picture
  quality.
\newblock \emph{IEEE TIP}, 2016.

\bibitem[Gu et~al.(2022)Gu, Wang, Xie, Dong, Li, Shan, and Cheng]{gu2022vqfr}
Yuchao Gu, Xintao Wang, Liangbin Xie, Chao Dong, Gen Li, Ying Shan, and
  Ming-Ming Cheng.
\newblock {VQFR}: Blind face restoration with vector-quantized dictionary and
  parallel decoder.
\newblock In \emph{ECCV}, 2022.

\bibitem[Han et~al.(2025)Han, Fan, Kong, Liao, Guo, Li, Timofte,
  et~al.]{ntire2025text}
Shuhao Han, Haotian Fan, Fangyuan Kong, Wenjie Liao, Chunle Guo, Chongyi Li,
  Radu Timofte, et~al.
\newblock {NTIRE} 2025 challenge on text to image generation model quality
  assessment.
\newblock In \emph{Proceedings of the IEEE/CVF Conference on Computer Vision
  and Pattern Recognition (CVPR) Workshops}, 2025.

\bibitem[Hosu et~al.(2020)Hosu, Lin, Sziranyi, and Saupe]{hosu2020koniq}
Vlad Hosu, Hanhe Lin, Tamas Sziranyi, and Dietmar Saupe.
\newblock {KonIQ-10k}: An ecologically valid database for deep learning of
  blind image quality assessment.
\newblock \emph{IEEE TIP}, 2020.

\bibitem[Huang et~al.(2008)Huang, Mattar, Berg, and
  Learned-Miller]{huang2008lfw}
Gary~B. Huang, Marwan Mattar, Tamara Berg, and Eric Learned-Miller.
\newblock {Labeled Faces in the Wild: A Database forStudying Face Recognition
  in Unconstrained Environments}.
\newblock In \emph{{Workshop on Faces in 'Real-Life' Images: Detection,
  Alignment, and Recognition}}, 2008.

\bibitem[Jain et~al.(2025)Jain, Wu, Zou, Florentin, Turbell, Siddhartha,
  Timofte, et~al.]{ntire2025vqe}
Varun Jain, Zongwei Wu, Quan Zou, Louis Florentin, Henrik Turbell, Sandeep
  Siddhartha, Radu Timofte, et~al.
\newblock {NTIRE} 2025 challenge on video quality enhancement for video
  conferencing: Datasets, methods and results.
\newblock In \emph{Proceedings of the IEEE/CVF Conference on Computer Vision
  and Pattern Recognition (CVPR) Workshops}, 2025.

\bibitem[Karras et~al.(2018)Karras, Aila, Laine, and
  Lehtinen]{karras2018celeba}
Tero Karras, Timo Aila, Samuli Laine, and Jaakko Lehtinen.
\newblock Progressive growing of {GAN}s for improved quality, stability, and
  variation.
\newblock In \emph{ICLR}, 2018.

\bibitem[Karras et~al.(2019)Karras, Laine, and Aila]{karras2019ffhq}
Tero Karras, Samuli Laine, and Timo Aila.
\newblock A style-based generator architecture for generative adversarial
  networks.
\newblock In \emph{CVPR}, 2019.

\bibitem[Ke et~al.(2021)Ke, Wang, Wang, Milanfar, and Yang]{ke2021musiq}
Junjie Ke, Qifei Wang, Yilin Wang, Peyman Milanfar, and Feng Yang.
\newblock { MUSIQ: Multi-scale Image Quality Transformer }.
\newblock In \emph{ICCV}, 2021.

\bibitem[Kim et~al.(2019)Kim, Kim, Kwon, and Kim]{kim2019progressive}
Deokyun Kim, Minseon Kim, Gihyun Kwon, and Dae-Shik Kim.
\newblock Progressive face super-resolution via attention to facial landmark.
\newblock In \emph{BMVC}, 2019.

\bibitem[Kim et~al.(2022)Kim, Jain, and Liu]{kim2022adaface}
Minchul Kim, Anil~K Jain, and Xiaoming Liu.
\newblock Adaface: Quality adaptive margin for face recognition.
\newblock In \emph{CVPR}, 2022.

\bibitem[Larson and Chandler(2010)]{larson2010most}
Eric~C. Larson and Damon~M. Chandler.
\newblock Most apparent distortion: Full-reference image quality assessment and
  the role of strategy.
\newblock \emph{JEI}, 2010.

\bibitem[Lee et~al.(2025)Lee, Park, Canelo, Park, Kim, Chun, Jin, Li, Guo,
  Timofte, et~al.]{ntire2025ebhdr}
Sangmin Lee, Eunpil Park, Angel Canelo, Hyunhee Park, Youngjo Kim, Hyungju
  Chun, Xin Jin, Chongyi Li, Chun-Le Guo, Radu Timofte, et~al.
\newblock {NTIRE} 2025 challenge on efficient burst hdr and restoration:
  Datasets, methods, and results.
\newblock In \emph{Proceedings of the IEEE/CVF Conference on Computer Vision
  and Pattern Recognition (CVPR) Workshops}, 2025.

\bibitem[Li et~al.(2025{\natexlab{a}})Li, Jin, Jin, Wu, Li, Wang, Yang, Li,
  Chen, Wen, Tan, Timofte, et~al.]{ntire2025day}
Xin Li, Yeying Jin, Xin Jin, Zongwei Wu, Bingchen Li, Yufei Wang, Wenhan Yang,
  Yu Li, Zhibo Chen, Bihan Wen, Robby Tan, Radu Timofte, et~al.
\newblock {NTIRE} 2025 challenge on day and night raindrop removal for
  dual-focused images: Methods and results.
\newblock In \emph{Proceedings of the IEEE/CVF Conference on Computer Vision
  and Pattern Recognition (CVPR) Workshops}, 2025{\natexlab{a}}.

\bibitem[Li et~al.(2025{\natexlab{b}})Li, Wang, Li, Yuan, Shao, Yao, Sun, Zhou,
  Timofte, and Chen]{ntire2025shortugc_data}
Xin Li, Xijun Wang, Bingchen Li, Kun Yuan, Yizhen Shao, Suhang Yao, Ming Sun,
  Chao Zhou, Radu Timofte, and Zhibo Chen.
\newblock {NTIRE} 2025 challenge on short-form ugc video quality assessment and
  enhancement: Kwaisr dataset and study.
\newblock In \emph{Proceedings of the IEEE/CVF Conference on Computer Vision
  and Pattern Recognition (CVPR) Workshops}, 2025{\natexlab{b}}.

\bibitem[Li et~al.(2025{\natexlab{c}})Li, Yuan, Li, Guan, Shao, Yu, Wang, Lu,
  Luo, Yao, Sun, Zhou, Chen, Timofte, et~al.]{ntire2025shortugc}
Xin Li, Kun Yuan, Bingchen Li, Fengbin Guan, Yizhen Shao, Zihao Yu, Xijun Wang,
  Yiting Lu, Wei Luo, Suhang Yao, Ming Sun, Chao Zhou, Zhibo Chen, Radu
  Timofte, et~al.
\newblock {NTIRE} 2025 challenge on short-form ugc video quality assessment and
  enhancement: Methods and results.
\newblock In \emph{Proceedings of the IEEE/CVF Conference on Computer Vision
  and Pattern Recognition (CVPR) Workshops}, 2025{\natexlab{c}}.

\bibitem[Li et~al.(2023)Li, Zhang, Liang, Cao, Liu, Gong, Zhang, Tang, Liu,
  Demandolx, Ranjan, Timofte, and Van~Gool]{li2023lsdir}
Yawei Li, Kai Zhang, Jingyun Liang, Jiezhang Cao, Ce Liu, Rui Gong, Yulun
  Zhang, Hao Tang, Yun Liu, Denis Demandolx, Rakesh Ranjan, Radu Timofte, and
  Luc Van~Gool.
\newblock Lsdir: A large scale dataset for image restoration.
\newblock In \emph{CVPRW}, 2023.

\bibitem[Liang et~al.(2021)Liang, Cao, Sun, Zhang, Van~Gool, and
  Timofte]{liang2021swinir}
Jingyun Liang, Jiezhang Cao, Guolei Sun, Kai Zhang, Luc Van~Gool, and Radu
  Timofte.
\newblock Swinir: Image restoration using swin transformer.
\newblock In \emph{ICCVW}, 2021.

\bibitem[Liang et~al.(2025)Liang, Timofte, Yi, Zhang, Liu, Sun, Wu, Zhang,
  Zeng, Zhang, et~al.]{ntire2025raim}
Jie Liang, Radu Timofte, Qiaosi Yi, Zhengqiang Zhang, Shuaizheng Liu, Lingchen
  Sun, Rongyuan Wu, Xindong Zhang, Hui Zeng, Lei Zhang, et~al.
\newblock {NTIRE} 2025 the 2nd restore any image model {(RAIM)} in the wild
  challenge.
\newblock In \emph{Proceedings of the IEEE/CVF Conference on Computer Vision
  and Pattern Recognition (CVPR) Workshops}, 2025.

\bibitem[Lin et~al.(2019)Lin, Hosu, and Saupe]{lin2019kadid}
Hanhe Lin, Vlad Hosu, and Dietmar Saupe.
\newblock {KADID-10k}: A large-scale artificially distorted {IQA} database.
\newblock In \emph{ICME}, 2019.

\bibitem[Lin et~al.(2024)Lin, He, Chen, Lyu, Dai, Yu, Ouyang, Qiao, and
  Dong]{lin2024diffbir}
Xinqi Lin, Jingwen He, Ziyan Chen, Zhaoyang Lyu, Bo Dai, Fanghua Yu, Wanli
  Ouyang, Yu Qiao, and Chao Dong.
\newblock {DiffBIR}: Towards blind image restoration with generative diffusion
  prior.
\newblock In \emph{ECCV}, 2024.

\bibitem[Liu et~al.(2025{\natexlab{a}})Liu, Min, Hu, Zhang, Guo,
  et~al.]{ntire2025xgc}
Xiaohong Liu, Xiongkuo Min, Qiang Hu, Xiaoyun Zhang, Jie Guo, et~al.
\newblock {NTIRE} 2025 {XGC} quality assessment challenge: Methods and results.
\newblock In \emph{Proceedings of the IEEE/CVF Conference on Computer Vision
  and Pattern Recognition (CVPR) Workshops}, 2025{\natexlab{a}}.

\bibitem[Liu et~al.(2025{\natexlab{b}})Liu, Wu, Vasluianu, Yan, Ren, Zhang, Gu,
  Zhang, Zhu, Timofte, et~al.]{ntire2025lowlight}
Xiaoning Liu, Zongwei Wu, Florin-Alexandru Vasluianu, Hailong Yan, Bin Ren,
  Yulun Zhang, Shuhang Gu, Le Zhang, Ce Zhu, Radu Timofte, et~al.
\newblock {NTIRE} 2025 challenge on low light image enhancement: Methods and
  results.
\newblock In \emph{Proceedings of the IEEE/CVF Conference on Computer Vision
  and Pattern Recognition (CVPR) Workshops}, 2025{\natexlab{b}}.

\bibitem[Miao et~al.(2024)Miao, Deng, and Han]{miao2024waveface}
Yunqi Miao, Jiankang Deng, and Jungong Han.
\newblock Waveface: Authentic face restoration with efficient frequency
  recovery.
\newblock In \emph{CVPR}, 2024.

\bibitem[Oquab et~al.(2023)Oquab, Darcet, Moutakanni, Vo, Szafraniec, Khalidov,
  Fernandez, Haziza, Massa, El-Nouby, et~al.]{oquab2023dinov2}
Maxime Oquab, Timoth{\'e}e Darcet, Th{\'e}o Moutakanni, Huy Vo, Marc
  Szafraniec, Vasil Khalidov, Pierre Fernandez, Daniel Haziza, Francisco Massa,
  Alaaeldin El-Nouby, et~al.
\newblock Dinov2: Learning robust visual features without supervision.
\newblock \emph{arXiv preprint arXiv:2304.07193}, 2023.

\bibitem[Podell et~al.(2023)Podell, English, Lacey, Blattmann, Dockhorn,
  M{\"u}ller, Penna, and Rombach]{SDXL}
Dustin Podell, Zion English, Kyle Lacey, Andreas Blattmann, Tim Dockhorn, Jonas
  M{\"u}ller, Joe Penna, and Robin Rombach.
\newblock Sdxl: Improving latent diffusion models for high-resolution image
  synthesis.
\newblock \emph{arXiv preprint arXiv:2307.01952}, 2023.

\bibitem[Qin et~al.(2019)Qin, Wang, Bai, Xie, and
  Jia]{qin2019ffanetfeaturefusionattention}
Xu Qin, Zhilin Wang, Yuanchao Bai, Xiaodong Xie, and Huizhu Jia.
\newblock Ffa-net: Feature fusion attention network for single image dehazing.
\newblock \emph{arXiv preprint arXiv:1911.07559}, 2019.

\bibitem[Qiu et~al.(2023)Qiu, Han, Zhang, Li, Guo, and Nie]{qiu2023diffbfr}
Xinmin Qiu, Congying Han, Zicheng Zhang, Bonan Li, Tiande Guo, and Xuecheng
  Nie.
\newblock Diffbfr: Bootstrapping diffusion model for blind face restoration.
\newblock In \emph{ACM MM}, 2023.

\bibitem[Ren et~al.(2025)Ren, Guo, Sun, Wu, Timofte, Li, et~al.]{ntire2025esr}
Bin Ren, Hang Guo, Lei Sun, Zongwei Wu, Radu Timofte, Yawei Li, et~al.
\newblock The tenth {NTIRE} 2025 efficient super-resolution challenge report.
\newblock In \emph{Proceedings of the IEEE/CVF Conference on Computer Vision
  and Pattern Recognition (CVPR) Workshops}, 2025.

\bibitem[Safonov et~al.(2025)Safonov, Bryntsev, Moskalenko, Kulikov, Vatolin,
  Timofte, et~al.]{ntire2025ugc}
Nickolay Safonov, Alexey Bryntsev, Andrey Moskalenko, Dmitry Kulikov, Dmitriy
  Vatolin, Radu Timofte, et~al.
\newblock {NTIRE} 2025 challenge on {UGC} video enhancement: Methods and
  results.
\newblock In \emph{Proceedings of the IEEE/CVF Conference on Computer Vision
  and Pattern Recognition (CVPR) Workshops}, 2025.

\bibitem[Sauer et~al.(2024)Sauer, Lorenz, Blattmann, and Rombach]{SDXL-Turbo}
Axel Sauer, Dominik Lorenz, Andreas Blattmann, and Robin Rombach.
\newblock Adversarial diffusion distillation.
\newblock In \emph{ECCV}, 2024.

\bibitem[Sheikh et~al.(2006)Sheikh, Sabir, and Bovik]{sheikh2006statistical}
Hamid~R. Sheikh, Muhammad~F. Sabir, and Alan~C. Bovik.
\newblock A statistical evaluation of recent full reference image quality
  assessment algorithms.
\newblock \emph{IEEE TIP}, 2006.

\bibitem[Shen et~al.(2018)Shen, Lai, Xu, Kautz, and Yang]{shen2018deep}
Ziyi Shen, Wei-Sheng Lai, Tingfa Xu, Jan Kautz, and Ming-Hsuan Yang.
\newblock Deep semantic face deblurring.
\newblock In \emph{CVPR}, 2018.

\bibitem[Shocher et~al.(2018)Shocher, Cohen, and Irani]{shocher2018zero}
Assaf Shocher, Nadav Cohen, and Michal Irani.
\newblock “zero-shot” super-resolution using deep internal learning.
\newblock In \emph{CVPR}, 2018.

\bibitem[Simonyan and Zisserman(2015)]{simonyan2015vgg}
Karen Simonyan and Andrew Zisserman.
\newblock Very deep convolutional networks for large-scale image recognition.
\newblock In \emph{ICLR}, 2015.

\bibitem[Suin and Chellappa(2024)]{Suin2024CLRFace}
Maitreya Suin and Rama Chellappa.
\newblock {CLR-Face}: Conditional latent refinement for blind face restoration
  using score-based diffusion models.
\newblock In \emph{IJCAI}, 2024.

\bibitem[Sun et~al.(2024)Sun, Wu, Ma, Liu, Yi, and Zhang]{sun2024pixel}
Lingchen Sun, Rongyuan Wu, Zhiyuan Ma, Shuaizheng Liu, Qiaosi Yi, and Lei
  Zhang.
\newblock Pixel-level and semantic-level adjustable super-resolution: A
  dual-lora approach.
\newblock \emph{arXiv preprint arXiv:2412.03017}, 2024.

\bibitem[Sun et~al.(2025{\natexlab{a}})Sun, Alfarano, Duan, Su, Wang, Shi,
  Timofte, Paudel, Van~Gool, et~al.]{ntire2025event}
Lei Sun, Andrea Alfarano, Peiqi Duan, Shaolin Su, Kaiwei Wang, Boxin Shi, Radu
  Timofte, Danda~Pani Paudel, Luc Van~Gool, et~al.
\newblock {NTIRE} 2025 challenge on event-based image deblurring: Methods and
  results.
\newblock In \emph{Proceedings of the IEEE/CVF Conference on Computer Vision
  and Pattern Recognition (CVPR) Workshops}, 2025{\natexlab{a}}.

\bibitem[Sun et~al.(2025{\natexlab{b}})Sun, Guo, Ren, Van~Gool, Timofte, Li,
  et~al.]{ntire2025denoising}
Lei Sun, Hang Guo, Bin Ren, Luc Van~Gool, Radu Timofte, Yawei Li, et~al.
\newblock The tenth ntire 2025 image denoising challenge report.
\newblock In \emph{Proceedings of the IEEE/CVF Conference on Computer Vision
  and Pattern Recognition (CVPR) Workshops}, 2025{\natexlab{b}}.

\bibitem[Sun et~al.(2025{\natexlab{c}})Sun, Wu, Ma, Liu, Yi, and
  Zhang]{sun2024pisasr}
Lingchen Sun, Rongyuan Wu, Zhiyuan Ma, Shuaizheng Liu, Qiaosi Yi, and Lei
  Zhang.
\newblock Pixel-level and semantic-level adjustable super-resolution: A
  dual-lora approach.
\newblock In \emph{CVPR}, 2025{\natexlab{c}}.

\bibitem[Tao et~al.(2025)Tao, Gu, Zhang, Wang, and Cheng]{tao2025overcoming}
Keda Tao, Jinjin Gu, Yulun Zhang, Xiucheng Wang, and Nan Cheng.
\newblock Overcoming false illusions in real-world face restoration with
  multi-modal guided diffusion model.
\newblock In \emph{ICLR}, 2025.

\bibitem[Tsai et~al.(2024)Tsai, Liu, Qi, Chan, and Yang]{tsai2024daefr}
Yu-Ju Tsai, Yu-Lun Liu, Lu Qi, Kelvin~CK Chan, and Ming-Hsuan Yang.
\newblock Dual associated encoder for face restoration.
\newblock In \emph{ICLR}, 2024.

\bibitem[Vasluianu et~al.(2025{\natexlab{a}})Vasluianu, Seizinger, Zhou, Chen,
  Wu, Timofte, et~al.]{ntire2025shadow}
Florin-Alexandru Vasluianu, Tim Seizinger, Zhuyun Zhou, Cailian Chen, Zongwei
  Wu, Radu Timofte, et~al.
\newblock {NTIRE} 2025 image shadow removal challenge report.
\newblock In \emph{Proceedings of the IEEE/CVF Conference on Computer Vision
  and Pattern Recognition (CVPR) Workshops}, 2025{\natexlab{a}}.

\bibitem[Vasluianu et~al.(2025{\natexlab{b}})Vasluianu, Seizinger, Zhou, Wu,
  Timofte, et~al.]{ntire2025ambient}
Florin-Alexandru Vasluianu, Tim Seizinger, Zhuyun Zhou, Zongwei Wu, Radu
  Timofte, et~al.
\newblock {NTIRE} 2025 ambient lighting normalization challenge.
\newblock In \emph{Proceedings of the IEEE/CVF Conference on Computer Vision
  and Pattern Recognition (CVPR) Workshops}, 2025{\natexlab{b}}.

\bibitem[Wallace et~al.(2023)Wallace, Gokul, and Naik]{wallace2023edict}
Bram Wallace, Akash Gokul, and Nikhil Naik.
\newblock Edict: Exact diffusion inversion via coupled transformations.
\newblock In \emph{CVPR}, 2023.

\bibitem[Wang et~al.(2023{\natexlab{a}})Wang, Chan, and Loy]{wang2022clipiqa}
Jianyi Wang, Kelvin~CK Chan, and Chen~Change Loy.
\newblock Exploring clip for assessing the look and feel of images.
\newblock In \emph{AAAI}, 2023{\natexlab{a}}.

\bibitem[Wang et~al.(2024)Wang, Yue, Zhou, Chan, and Loy]{wang2024exploiting}
Jianyi Wang, Zongsheng Yue, Shangchen Zhou, Kelvin~C.K. Chan, and Chen~Change
  Loy.
\newblock Exploiting diffusion prior for real-world image super-resolution.
\newblock \emph{IJCV}, 2024.

\bibitem[Wang et~al.(2025{\natexlab{a}})Wang, Gong, Zhang, Chen, Liu, Gu, Liu,
  Zhang, and Yang]{wang2025osdface}
Jingkai Wang, Jue Gong, Lin Zhang, Zheng Chen, Xing Liu, Hong Gu, Yutong Liu,
  Yulun Zhang, and Xiaokang Yang.
\newblock One-step diffusion model for face restoration.
\newblock In \emph{CVPR}, 2025{\natexlab{a}}.

\bibitem[Wang et~al.(2021{\natexlab{a}})Wang, Li, Zhang, and
  Shan]{wang2021gfpgan}
Xintao Wang, Yu Li, Honglun Zhang, and Ying Shan.
\newblock Towards real-world blind face restoration with generative facial
  prior.
\newblock In \emph{CVPR}, 2021{\natexlab{a}}.

\bibitem[Wang et~al.(2021{\natexlab{b}})Wang, Xie, Dong, and
  Shan]{wang2021real}
Xintao Wang, Liangbin Xie, Chao Dong, and Ying Shan.
\newblock Real-esrgan: Training real-world blind super-resolution with pure
  synthetic data.
\newblock In \emph{ICCVW}, 2021{\natexlab{b}}.

\bibitem[Wang et~al.(2025{\natexlab{b}})Wang, Liang, Zhang, Tian, Wang, Li,
  Yang, Timofte, Guo, et~al.]{ntire2025lightfield}
Yingqian Wang, Zhengyu Liang, Fengyuan Zhang, Lvli Tian, Longguang Wang,
  Juncheng Li, Jungang Yang, Radu Timofte, Yulan Guo, et~al.
\newblock {NTIRE} 2025 challenge on light field image super-resolution: Methods
  and results.
\newblock In \emph{Proceedings of the IEEE/CVF Conference on Computer Vision
  and Pattern Recognition (CVPR) Workshops}, 2025{\natexlab{b}}.

\bibitem[Wang et~al.(2023{\natexlab{b}})Wang, Zhang, Chen, Wang, and
  Luo]{wang2023restoreformer++}
Zhouxia Wang, Jiawei Zhang, Tianshui Chen, Wenping Wang, and Ping Luo.
\newblock Restoreformer++: Towards real-world blind face restoration from
  undegraded key-value pairs.
\newblock \emph{IEEE TPAMI}, 2023{\natexlab{b}}.

\bibitem[Wang et~al.(2023{\natexlab{c}})Wang, Zhang, Zhang, Zheng, Zhou, Zhang,
  and Wang]{wang2023dr2}
Zhixin Wang, Xiaoyun Zhang, Ziying Zhang, Huangjie Zheng, Mingyuan Zhou, Ya
  Zhang, and Yanfeng Wang.
\newblock Dr2: Diffusion-based robust degradation remover for blind face
  restoration.
\newblock In \emph{CVPR}, 2023{\natexlab{c}}.

\bibitem[Wu et~al.(2024{\natexlab{a}})Wu, Zhang, Zhang, Chen, Li, Liao, Wang,
  Zhang, Sun, Yan, Min, Zhai, and Lin]{wu2024qalign}
Haoning Wu, Zicheng Zhang, Weixia Zhang, Chaofeng Chen, Chunyi Li, Liang Liao,
  Annan Wang, Erli Zhang, Wenxiu Sun, Qiong Yan, Xiongkuo Min, Guangtai Zhai,
  and Weisi Lin.
\newblock Q-align: Teaching lmms for visual scoring via discrete text-defined
  levels.
\newblock In \emph{ICML}, 2024{\natexlab{a}}.

\bibitem[Wu et~al.(2024{\natexlab{b}})Wu, Sun, Ma, and Zhang]{wu2024osediff}
Rongyuan Wu, Lingchen Sun, Zhiyuan Ma, and Lei Zhang.
\newblock One-step effective diffusion network for real-world image
  super-resolution.
\newblock In \emph{NeurIPS}, 2024{\natexlab{b}}.

\bibitem[Wu et~al.(2024{\natexlab{c}})Wu, Yang, Sun, Zhang, Li, and
  Zhang]{wu2024seesr}
Rongyuan Wu, Tao Yang, Lingchen Sun, Zhengqiang Zhang, Shuai Li, and Lei Zhang.
\newblock {SeeSR}: Towards semantics-aware real-world image super-resolution.
\newblock In \emph{CVPR}, 2024{\natexlab{c}}.

\bibitem[Xie et~al.(2024)Xie, Zheng, Xue, Jiang, Liu, Wu, and
  Wong]{xie2024pltrans}
Lianxin Xie, Csbingbing Zheng, Wen Xue, Le Jiang, Cheng Liu, Si Wu, and Hau~San
  Wong.
\newblock Learning degradation-unaware representation with prior-based latent
  transformations for blind face restoration.
\newblock In \emph{CVPR}, 2024.

\bibitem[Yang et~al.(2025)Yang, Cai, Ouyang, Vasluianu, Timofte, Ding, Sun, Fu,
  Li, Ho, Meng, et~al.]{ntire2025reflection}
Kangning Yang, Jie Cai, Ling Ouyang, Florin-Alexandru Vasluianu, Radu Timofte,
  Jiaming Ding, Huiming Sun, Lan Fu, Jinlong Li, Chiu~Man Ho, Zibo Meng, et~al.
\newblock {NTIRE} 2025 challenge on single image reflection removal in the
  wild: Datasets, methods and results.
\newblock In \emph{Proceedings of the IEEE/CVF Conference on Computer Vision
  and Pattern Recognition (CVPR) Workshops}, 2025.

\bibitem[Yang et~al.(2023)Yang, Zhou, Tao, and Loy]{yang2023pgdiff}
Peiqing Yang, Shangchen Zhou, Qingyi Tao, and Chen~Change Loy.
\newblock {PGDiff}: Guiding diffusion models for versatile face restoration via
  partial guidance.
\newblock In \emph{NeurIPS}, 2023.

\bibitem[Yang et~al.(2022)Yang, Wu, Shi, Lao, Gong, Cao, Wang, and
  Yang]{yang2022maniqa}
Sidi Yang, Tianhe Wu, Shuwei Shi, Shanshan Lao, Yuan Gong, Mingdeng Cao, Jiahao
  Wang, and Yujiu Yang.
\newblock {MANIQA}: Multi-dimension attention network for no-reference image
  quality assessment.
\newblock In \emph{CVPRW}, 2022.

\bibitem[Yang et~al.(2021)Yang, Ren, Xie, and Zhang]{Yang2021GPEN}
Tao Yang, Peiran Ren, Xuansong Xie, and Lei Zhang.
\newblock Gan prior embedded network for blind face restoration in the wild.
\newblock In \emph{CVPR}, 2021.

\bibitem[Yu et~al.(2024)Yu, Gu, Li, Hu, Kong, Wang, He, Qiao, and
  Dong]{yu2024scaling}
Fanghua Yu, Jinjin Gu, Zheyuan Li, Jinfan Hu, Xiangtao Kong, Xintao Wang,
  Jingwen He, Yu Qiao, and Chao Dong.
\newblock Scaling up to excellence: Practicing model scaling for
  photo-realistic image restoration in the wild.
\newblock In \emph{CVPR}, 2024.

\bibitem[Yu et~al.(2018)Yu, Fernando, Hartley, and Porikli]{yu2018super}
Xin Yu, Basura Fernando, Richard Hartley, and Fatih Porikli.
\newblock Super-resolving very low-resolution face images with supplementary
  attributes.
\newblock In \emph{CVPR}, 2018.

\bibitem[Yue and Loy(2024)]{yue2024difface}
Zongsheng Yue and Chen~Change Loy.
\newblock { DifFace: Blind Face Restoration with Diffused Error Contraction }.
\newblock \emph{IEEE TPAMI}, 2024.

\bibitem[Zama~Ramirez et~al.(2025)Zama~Ramirez, Tosi, Di~Stefano, Timofte,
  Costanzino, Poggi, Salti, Mattoccia, et~al.]{ntire2025hrdepth}
Pierluigi Zama~Ramirez, Fabio Tosi, Luigi Di~Stefano, Radu Timofte, Alex
  Costanzino, Matteo Poggi, Samuele Salti, Stefano Mattoccia, et~al.
\newblock {NTIRE} 2025 challenge on hr depth from images of specular and
  transparent surfaces.
\newblock In \emph{Proceedings of the IEEE/CVF Conference on Computer Vision
  and Pattern Recognition (CVPR) Workshops}, 2025.

\bibitem[Zhang et~al.(2024{\natexlab{a}})Zhang, Alaluf, Ma, Kadambi, Wang, and
  Aberman]{InstantRestore}
Howard Zhang, Yuval Alaluf, Sizhuo Ma, Achuta Kadambi, Jian Wang, and Kfir
  Aberman.
\newblock Instantrestore: Single-step personalized face restoration with
  shared-image attention.
\newblock \emph{arXiv preprint arXiv:2412.06753}, 2024{\natexlab{a}}.

\bibitem[Zhang et~al.(2015)Zhang, Zhang, and Bovik]{zhang2015niqe}
Lin Zhang, Lei Zhang, and Alan~C. Bovik.
\newblock A feature-enriched completely blind image quality evaluator.
\newblock \emph{IEEE TIP}, 2015.

\bibitem[Zhang et~al.(2023)Zhang, Zhai, Wei, Yang, and Ma]{zhang2023blind}
Weixia Zhang, Guangtao Zhai, Ying Wei, Xiaokang Yang, and Kede Ma.
\newblock Blind image quality assessment via vision-language correspondence: A
  multitask learning perspective.
\newblock In \emph{CVPR}, 2023.

\bibitem[Zhang et~al.(2024{\natexlab{b}})Zhang, Li, Zhai, Yang, and
  Ma]{zhang2024no}
Weixia Zhang, Dingquan Li, Guangtao Zhai, Xiaokang Yang, and Kede Ma.
\newblock When no-reference image quality models meet map estimation in
  diffusion latents.
\newblock \emph{arXiv preprint arXiv:2403.06406}, 2024{\natexlab{b}}.

\bibitem[Zhou et~al.(2022)Zhou, Chan, Li, and Loy]{zhou2022codeformer}
Shangchen Zhou, Kelvin~C.K. Chan, Chongyi Li, and Chen~Change Loy.
\newblock Towards robust blind face restoration with codebook lookup
  transformer.
\newblock In \emph{NeurIPS}, 2022.

\end{thebibliography}
}

\end{document}


\maketitle

\appendix
\section{More Challenge Methods and Teams}
\label{sec:teams}

\subsection{UpHorse}

\begin{figure*}[t]
    \centering
    \includegraphics[width=1\linewidth]{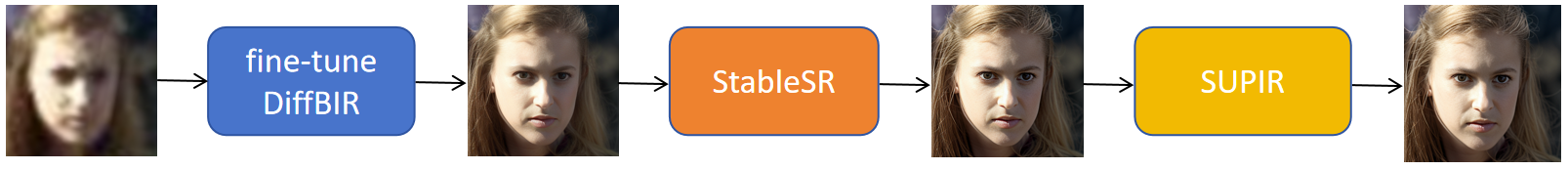}
    \caption{\textbf{Team UpHorse.} Overall pipeline. }
    \label{fig:Team14Fig4}
\end{figure*}

\begin{figure}[t]
    \centering
    \includegraphics[width=1\linewidth]{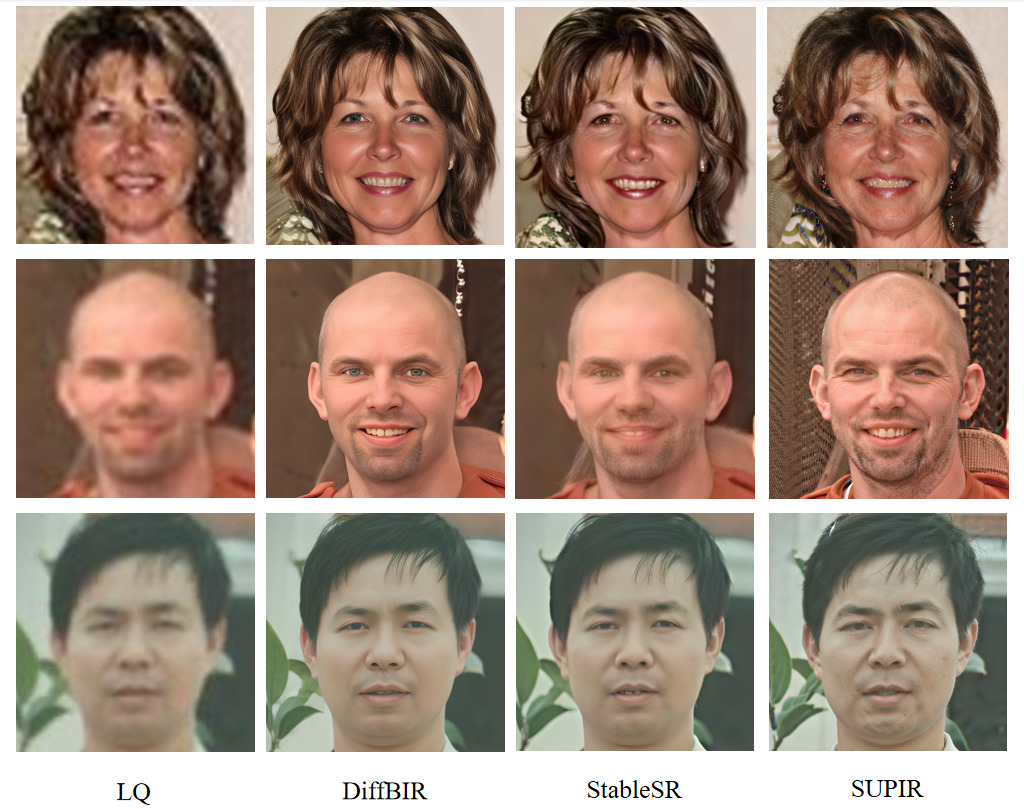}
    \caption{\textbf{Team UpHorse.} The results in initial restoration stage.}
    \label{fig:Team14Fig5}
\end{figure}

\begin{figure}[t]
    \centering
    \includegraphics[width=1\linewidth]{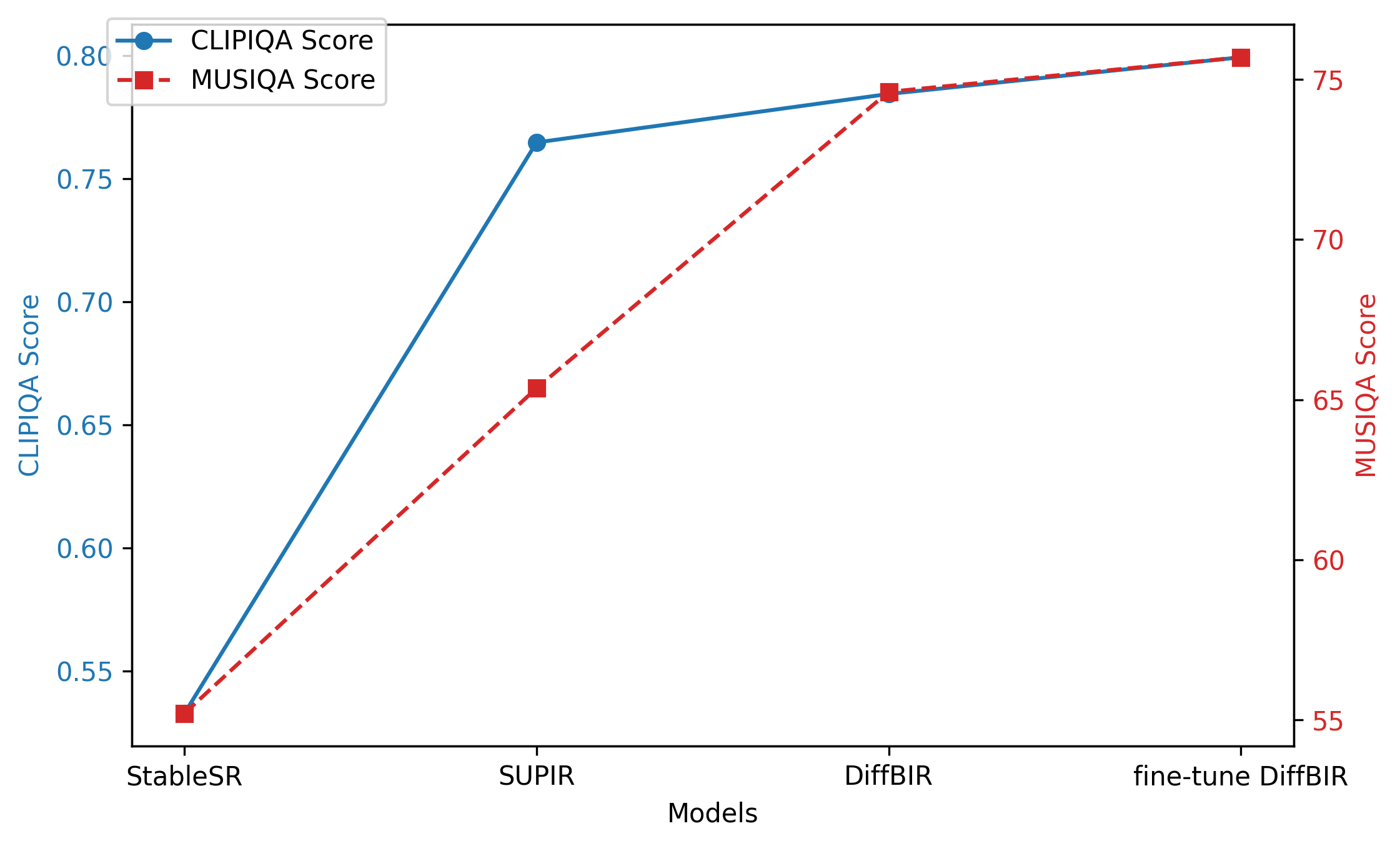}
    \caption{\textbf{Team UpHorse.} The scores in initial restoration stage.}
    \label{fig:Team14Fig6}
\end{figure}




\noindent\textbf{Description.} The proposed diffusion-prior-based implicit representation
joint face restoration method, DSS (Fig.~\ref{fig:Team14Fig4}), utilizes three
SOTA restoration and SR models for
the joint restoration of degraded images. Specifically, the
process begins with fine-tuning DiffBIR~\cite{lin2024diffbir} to recover the
fundamental structure of the image. Next, StableSR~\cite{wang2024exploiting} is
employed to precisely align the facial regions in real-world
scenarios. Finally, SUPIR~\cite{yu2024scaling} is used to further enhance
the facial details. By effectively leveraging the strengths
of each model, DSS achieves comprehensive, high-quality 
restoration of degraded images, ensuring superior performance across various aspects of the restoration process.

\begin{figure}[t]
    \centering
    \includegraphics[width=1\linewidth]{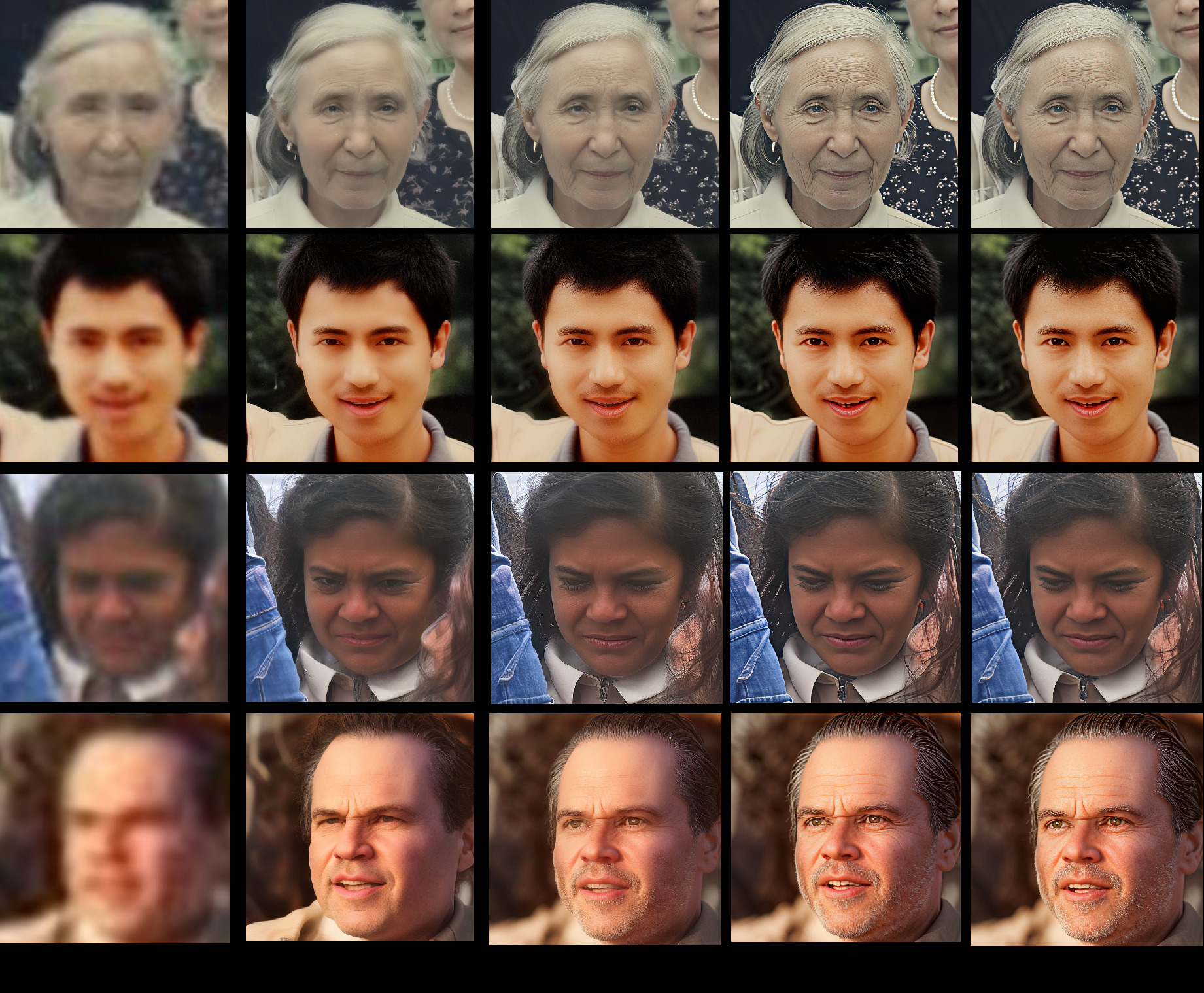}
    \caption{\textbf{Team UpHorse.} The restoration results at each stage. }
    \label{fig:Team14Fig7}
\end{figure}

\begin{figure}[t]
    \centering
    \includegraphics[width=1\linewidth]{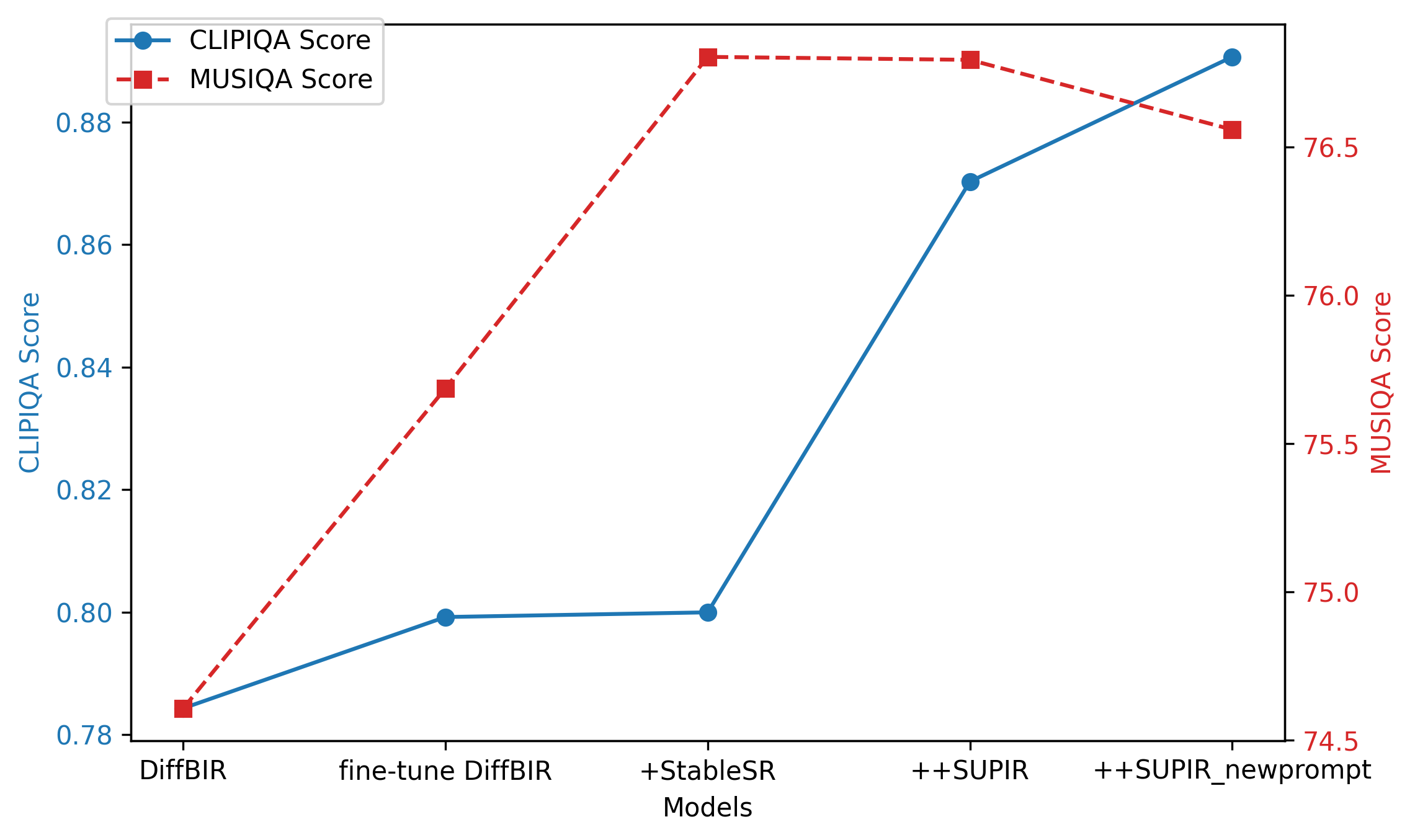}
    \caption{\textbf{Team UpHorse.} The scores at each restoration stage.}
    \label{fig:Team14Fig8}
\end{figure}

\begin{figure*}[t]
  \centering
  \includegraphics[width=\linewidth]{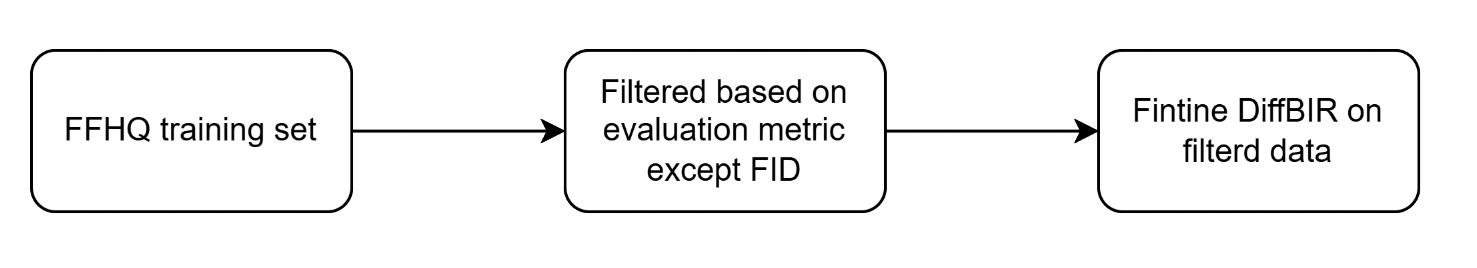}
  \caption{\textbf{Team CX.} }
  \label{fig:team11fig1}
\end{figure*}

\noindent\textbf{Implementation Details.} For images with varying degradation levels across different scenes, they apply three separate models for individual restoration. Experimental Fig.~\ref{fig:Team14Fig5} results show that, in the initial restoration phase, DiffBIR~\cite{lin2024diffbir} achieves superior results, while StableSR~\cite{wang2024exploiting} and SUPIR~\cite{yu2024scaling} tend to restore
more details. Based on these observations, they propose a three-stage face restoration framework: first, DiffBIR~\cite{lin2024diffbir}
is used for preliminary restoration to address the basic structure of the image; then, StableSR~\cite{wang2024exploiting} is employed to further align the facial features with real human faces; finally,
SUPIR~\cite{yu2024scaling} is applied to enhance facial details, thereby
achieving comprehensive and high-quality restoration.

To improve the degradation removal performance of
DiffBIR~\cite{lin2024diffbir} in the first stage, they fine-tune the first-stage
SwinIR~\cite{liang2021swinir} of DiffBIR~\cite{lin2024diffbir} using 50,000 FFHQ~\cite{karras2019ffhq} images. Specifically, they crop the input images to 512$\times$512, set the learning rate to 1e-4, and train for 150K iterations on an NVIDIA 4090 GPU. Considering the varying degradation
levels in the test data, they divided the degradation settings
into different ranges to better accommodate different levels
of image degradation. The loss function used is the Mean
Squared Error (MSE) loss Equation.~\ref{eq:team14eq1}. 
\begin{equation}
    I_{SW} = SW(I_{{lq}}), L = \Vert I_{SW}-I_{hq}\Vert^2_2,
    \label{eq:team14eq1}
\end{equation}
Where SW refers to SwinIR, $I_{SW}$ represents the output of
SwinIR, $I_{lq}$ denotes the low-quality image, and $I_{hq}$ denotes the
high-quality image.

In terms of the cfg scale parameter setting, larger values result in lower fidelity and more details. Given the input
data distribution requirements of subsequent models, they set
the cfg scale to 6 during the restoration process with DiffBIR~\cite{lin2024diffbir}, thus achieving a balance between the restoration
quality and the input requirements for subsequent detail enhancement. Other parameters remain unchanged from the
original DiffBIR~\cite{lin2024diffbir}. Experimental Fig.~\ref{fig:Team14Fig6} results demonstrate that the fine-tuned DiffBIR~\cite{lin2024diffbir} achieves significantly improved performance in degradation removal.

The experiments in Fig.~\ref{fig:Team14Fig7} and Fig.~\ref{fig:Team14Fig8} (``+" and ``++" indicate cumulative improvements over the previous method) show that in the second stage,
using StableSR~\cite{wang2024exploiting} on top of the first-stage restoration
achieves further alignment with real-world faces and enhances fine details. They kept the inference settings of StableSR~\cite{wang2024exploiting} unchanged. In the third stage, using SUPIR~\cite{yu2024scaling}
based on the second stage further enhances the texture details of the face.

Given SUPIR’s~\cite{yu2024scaling} powerful super-resolution capabilities and considering limited computational resources, they
removed the language-guided restoration module LLaVA~\cite{liu2024improvedbaselinesvisualinstruction}. Instead, they upsampled the input images by a factor of 3 and then downsampled them to create relatively
low-quality input images, which are specifically targeted by
SUPIR~\cite{yu2024scaling}. This adjustment led to an improvement in the
results after using SUPIR~\cite{yu2024scaling} for restoration. Additionally,
they experimented with various forward and reverse prompt
combinations to replace the original prompt, leading to further breakthroughs in the results.

\begin{figure*}[t]
    \centering
    \includegraphics[width=0.8\textwidth]{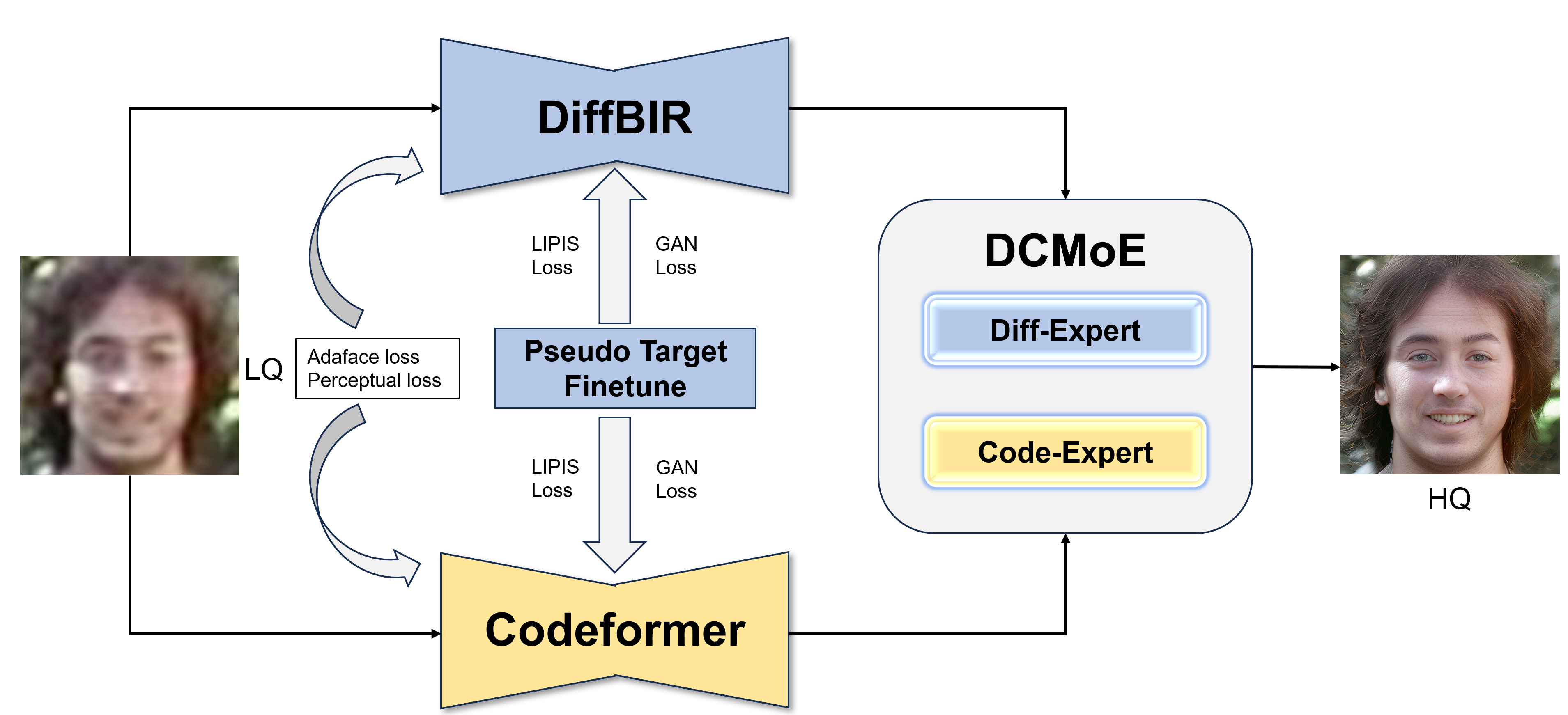}
    \caption{\textbf{Team AIIALab.} }
    \label{fig:team16-pipeline}
\end{figure*}

\subsection{CX}
\noindent\textbf{Description.} Their network design, shown in Fig.~\ref{fig:team11fig1}, is exactly the same as DiffBIR. They only performed data filtering and processing on FFHQ and then fine-tuned based on the pre-trained model of DiffBIR. 

\noindent\textbf{Implementation Details.} The proposed method achieves a runtime of 5 seconds per image (512x512) on an NVIDIA RTX 4090. It utilizes the DiffBIR model (version: v1\_face) and is trained exclusively on the FFHQ dataset. The training process uses Adam as the optimizer with a learning rate of 1.17e-6 for 80k iterations, conducted on a 512×512 resolution using 5 NVIDIA A5000 GPUs with a batch size of 30 (6 per GPU). During testing, 50 steps are performed with an empty positive prompt and a negative prompt of ``low quality, blurry, low-resolution, noisy, unsharp, weird textures", without using any image captioner for guidance. The method shows a 0.06 improvement in the final score: the weighted score of original v1\_face DiffBIR is 3.97, after finetuning on filtered FFHQ, the weighted score is 4.03. The solution is based on filtering the FFHQ training set.

\subsection{AIIALab}

\noindent\textbf{Description.} They address the challenge of restoring high-quality facial images from low-quality real-world inputs while preserving ID consistency. To achieve high-fidelity reconstruction with enhanced visual quality, they propose a dual-model adaptive Mixture-of-Experts (MoE) architecture that synergistically integrates the high-quality generative priors of diffusion models with the identity-preserving capabilities of transformer networks. Specifically, their framework combines the complementary strengths of DiffBIR (for quality enhancement through diffusion processes) and CodeFormer (for structural fidelity via transformer-based facial priors), implementing dynamic feature fusion through their novel adaptive gating mechanism.

\noindent\textbf{Implementation Details.}
DiffBIR\cite{lin2024diffbir} proposes a two-stage framework for blind image restoration that systematically decouples degradation removal and content reconstruction. In the first stage, they employ dedicated restoration modules to eliminate image-independent degradations (e.g., noise, blur), leveraging existing or custom-trained models (e.g., swinir) tailored for specific distortion types. The second stage focuses exclusively on semantic-aware content regeneration through their designed generation module, which operates solely on the purified image content from the first stage, thereby avoiding interference from residual artifacts. Crucially, they implement independent optimization strategies for both stages while maintaining inter-stage compatibility. To achieve dynamic quality control, they further develop a training-free adaptive guidance mechanism that spatially modulates restoration intensity during the diffusion sampling process, enabling a precise balance between perceptual quality and structural fidelity across different image regions. This architecture provides a unified yet flexible solution that combines task adaptability with stable performance across diverse blind restoration scenarios.

Codeformer\cite{zhou2022codeformer} begins by integrating vector quantization principles to establish a semantic-aware codebook through a self-reconstruction-driven pre-training process, where they first train a quantized autoencoder to learn discrete latent representations and their corresponding decoder. Building upon this learned codebook prior, they then design a Transformer-based architecture to precisely predict optimal code combinations directly from degraded facial inputs, enabling targeted restoration of missing facial details. To dynamically balance perceptual quality and identity preservation, they further develop a controllable feature transformation module that adaptively adjusts feature representations during the restoration process. The entire system follows a three-stage progressive training strategy: codebook construction, code prediction optimization, and fidelity-quality adaptive refinement, ensuring systematic alignment between prior knowledge extraction and task-specific restoration objectives.

Their method's pipeline is shown in Fig.~\ref{fig:team16-pipeline}. Their framework introduces an adaptive Mixture-of-Experts (MoE) module to dynamically integrate the outputs of DiffBIR and Codeformer, optimizing the selection of high-fidelity and visually plausible restored facial images. Specifically, the MoE module operates through three key mechanisms. \textbf{1)} The Diffusion Expert leverages iterative denoising to recover fine-grained details but may introduce identity shifts under severe degradations, while the Codebook Expert utilizes a pretrained vector-quantized autoencoder to enforce identity consistency through discrete code prediction, albeit with limited detail recovery capability. \textbf{2)} A lightweight gating network analyzes multi-scale degradation features (e.g., noise distribution, blur kernels) and semantic cues (facial landmarks, identity embeddings) from the low-quality input. It predicts dynamic weights $w_\text{diff}$ and $w_\text{code} = 1 - w_\text{diff}$
via a Sigmoid-activated MLP, prioritizing DiffBIR for noise/blur-dominated inputs and Codeformer for identity-critical scenarios (e.g., extreme low-resolution or occlusions). \textbf{3)} The final output is computed as \( I_{\text{final}} = w_{\text{diff}} \cdot I_{\text{diff}} + w_{\text{code}} \cdot I_{\text{code}} \), jointly optimized by a hybrid loss combining pixel-level reconstruction (\( \mathcal{L}_{\text{L1}} \)), identity preservation (\( \mathcal{L}_{\text{ID}} \)), and adversarial training (\( \mathcal{L}_{\text{adv}} \)). 

\begin{figure*}[t]
    \centering
    \includegraphics[width=1\linewidth]{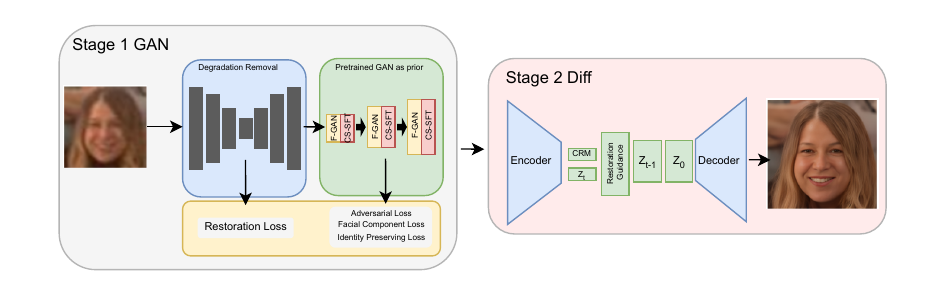}
    \caption{\textbf{Team ACVLab.}}
    \label{fig:team08-enter-label}
\end{figure*}

\begin{figure*}[t]
	\centering
        \footnotesize
	\includegraphics[width=1.0\linewidth]{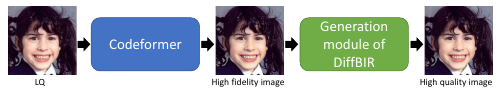}
    \caption{\textbf{Team HH.}}
    \label{fig:team04-network}
\end{figure*}

\noindent\textit{Training datasets.}
They use 70k FFHQ as their train datasets, no other data. All images are randomly cropped to 512 × 512 during training.
And they used data augmentation methods such as random rotation and flipping to expand the diversity of the dataset.

\noindent\textit{Training strategy.}
In the DiffBIR~\cite{lin2024diffbir}, they train the restoration module for 150k iterations (batch size=96). Then, they adopt Stable Diffusion 2.1-base1 as the generative prior, and finetune the proposed IRControlNet for 80k iterations (batch size=256). Adam is used as the optimizer. The learning rate is set to $10^{-4}$ for the first 30k iterations and then decreased to $10^{-5}$ for the following 50k iterations. 

In the codeformer~\cite{zhou2022codeformer}, they represent a face image of 512 × 512 as a 16 × 16 code sequence. For all stages of training, they use the Adam optimizer with a batch size of 16. They set the learning rate to 8×$10^{-5}$ for stages I and II, and adopt a smaller learning rate of 2×$10^{-5}$ for stage III. The three stages are trained with 1.5M, 200K, and 20K iterations, respectively. 

In the training process of the above two models, they also added perceptual loss and AdaFace loss to obtain high-fidelity and high-quality facial images, with weights of 0.01 for both losses.

After that, they also apply the DT-BFR\cite{kuai2024towards} method to finetune the SwinIR model and Codeformer model (SwinIR model is used as the DiffBIR stage 1 model). They first generate pseudo targets using a diffusion model and then use the generated targets to fine-tune the two pre-trained restoration models. For fine-tuning the SwinIR model, they set the weights of the losses to be $\lambda_\text{LPIPS} = 0.1$ and $\lambda_\text{GAN} = 0.1$ for all the experiments. For CodeFormer, they follow their training setup and empirically found that only adopting their code-level losses to optimize the code prediction module and the VQ-GAN encoder gives better fine-tuning performance than the image-level losses.

\subsection{ACVLab}

\noindent\textbf{Description.} As Fig.~\ref{fig:team08-enter-label}, they proposed a two-stage restoration approach for blind face restoration that combined GFPGAN \cite{wang2021gfpgan} and DiffBIR \cite{lin2024diffbir} to enhance real-world degraded facial images. GFPGAN is applied to the cropped face images to perform coarse restoration. This step effectively reconstructs missing facial details and provides an initial enhancement of the global structure while maintaining identity consistency. Afterwards, the output of GFPGAN is fed into DiffBIR, a diffusion-based blind image restoration model for fine-grained-level restoration and refinement.

\noindent\textbf{Implementation Details.} The training process utilizes the FFHQ \cite{karras2019ffhq} and FFHQR \cite{Shafaei2021ffhqr} datasets at a resolution of 512 $\times$ 512 pixels. They performed image degradation realistically via the standard degradation pipeline offered by the organizer on Codalab. The degraded input images undergo initial alignment and normalization to ensure consistency across different samples. RetinaFace \cite{Deng2020RetinaFace} is applied to locate facial regions, which are then cropped and resized to a standard resolution.

For training GFPGAN, their training utilizes several key optimization parameters. Adam optimizer \cite{kingma2014adam} is employed for both generator and discriminator networks with a learning rate of $2\times10^{-3}$. A MultiStepLR scheduler is implemented with milestones at 600,000 and 700,000 iterations (gamma 0.5). The total training process consists of 800,000 iterations. The discriminator is trained at a frequency of once per generator update. As for the objective function, they follow standard setting from the official repository of GFPGAN.

\subsection{HH}

\noindent\textbf{Description.} Their method (Fig.~\ref{fig:team04-network}) relies on DiffBIR~\cite{lin2024diffbir} and CodeFormer~\cite{zhou2022codeformer}. DiffBIR is a two-stage image restoration model. In the first stage, a base restoration model is responsible for generating a high-fidelity preliminary result. In the second stage, a diffusion model based on ControlNet conditional control is employed to add high-frequency details to the output of the first stage. they utilize CodeFormer as the first-stage model of DiffBIR, combining it with DiffBIR's second-stage model to achieve high-fidelity and high-quality portrait restoration.

\noindent\textbf{Implementation Details.} Their method integrates \textit{CodeFormer} and \textit{DiffBIR}, two existing approaches, and thus does not require retraining. The training strategies for these methods can be referenced in their respective papers.

\subsection{Fustar-fsr}
\noindent\textbf{Description.} They present their model named Incremental Face Restoration Model. This model has a hierarchical face restoration
framework, which integrates progressive generative diffusion models and face prior guidance, and is tailored to address the complex degradation issues in real-world scenarios. By leveraging the unique capabilities of PGDiff and GFPGAN, and incorporating post-stage image enhancement, they aim to achieve superior restoration results.

\noindent\textbf{Implementation Details.} Many current blind face restoration methods are quite excellent, which is beneficial for us to refer to them and achieve real-world face super-resolution. For instance, methods based on Generative Adversarial Networks (GANs), like GFPGAN (Generative Facial Prior-GAN), leverage the rich and diverse prior knowledge contained in pre-trained face GANs (such as StyleGAN2). This prior knowledge encompasses various features and structural information of the face, including the distribution of facial features, texture patterns, and colors, to guide the face restoration process. In the context of real-world face super-resolution, they can draw on this prior knowledge to assist the model in better understanding the structure and features of the face, thereby generating more reasonable and natural details when upscaling the image. For example, DiffBIR: It decomposes the blind image restoration problem into two sub-problems: degradation removal and detail generation. A two-stage framework is proposed. In the first stage, the restoration module trained by the MSE loss function is used to remove most of the degradation contained in the image, obtaining a clean but smooth image lacking local texture details. In the second stage, the ControlNet module is used to leverage the generative power of Stable Diffusion to compensate for lost texture details or semantic information. Meanwhile, a controllable module without the need for additional training-latent image guidance, is introduced to balance image quality and fidelity. And CodeFormer: It transforms the blind face restoration problem into a code prediction task. A discrete codebook is used to represent the local features of high-quality face images, and low-quality face images are mapped to the code space of high-quality images. By introducing the codebook lookup Transformer (CLT), it combines the advantages of discrete codebooks and Transformers. The self-attention mechanism is adopted to model the global and local features in images. And a hybrid training strategy is used to combine reconstruction error and perceptual loss to ensure that the model can not only remove noise and blur but also retain the realism and naturalness of the image during the restoration process.

After referring to those models, they primarily propose a hierarchical face restoration framework based on a progressive generative diffusion model and face prior guidance, and conduct further image enhancement in the subsequent stage. As the first-step processing method, they present a multi-stage cascaded framework. Specifically, this method successfully integrates the temporally controllable generation ability of the progressive generative diffusion model (PGDiff) with the structural semantic prior of the face prior generation model (GFPGAN), achieving progressive face restoration from coarse-grained to fine-grained levels.

\noindent\textit{(1) Utilization of PGDiff’s multi-scale diffusion guidance.}
They make use of PGDiff’s conditional-guided diffusion mechanism. During the diffusion process, they introduce time-step-adaptive semantic constraints. By dynamically fusing low-quality inputs (LQ) with pre-generated face prior features of GFPGAN at different diffusion stages (ranging from the high-noise level where $t = T$ to the low-noise level where $t = 0$), they realize progressive restoration from the global structure to local details.

\noindent\textit{(2) Semantic prior embedding of GFPGAN}
They regard GFPGAN as a semantic prior generator. Through adaptive instance normalization (AdaIN), it encodes its pre-trained knowledge of face structure (such as the distribution of facial features and texture patterns) into conditional vectors. These vectors are then input as class-conditional (class-cond) information during the reverse sampling process of PGDiff, constraining the diffusion model to generate high-resolution images that conform to the statistical characteristics of real faces.

\noindent\textit{(3) Subsequent-stage image enhancement processing}
To obtain more vivid images in terms of visual perception, they draw on the methods in the HGGT paper and design an image enhancement model aiming to further enhance the perceptual quality of the restored face images. Specifically, they train an SR model named ELAN with L1 loss, perceptual loss, and adversarial loss, and apply it to HQ images to further improve the perceptual quality of these images. They use the commonly used DF2K as the training set and degrade the HR images. Before inputting the degraded images into the model training, they resize them to the size of HR images. Additionally, they remove the upsampling layer in the model because, in order to obtain face images with enhanced perceptual quality, the input and output need to be the same size.

\noindent\textit{Training dataset.}
FFHQ is a high-quality image dataset of human faces, originally created as a benchmark for generative adversarial networks (GAN). The dataset consists of 70,000 high-quality PNG images at 1024×1024 resolution and contains considerable variation in terms of age, ethnicity, and image background. It also has good coverage of accessories such as eyeglasses, sunglasses, hats, etc. The images were crawled from Flickr, thus inheriting all the biases of that website, and automatically aligned and cropped using dlib. Only images under permissive licenses were collected. Various automatic filters were used to prune the set, and finally, Amazon Mechanical Turk was used to remove the occasional statues, paintings, or photos of photos. The participants are allowed to use extra data for training.

\noindent\textit{Training strategy.}
They use the Adam optimizer for training their models. The optimizer’s parameters are set as follows: $\beta_1 = 0.9$, $\beta_2 = 0.999$, and $\epsilon = 1e^{-8}$. These values have been empirically found to provide stable convergence during training.

\subsection{Night Watch}
\noindent\textbf{Description.} They use the initial weights provided by \url{https://github.com/XPixelGroup/DiffBIR}. Specifically, the original SwinIR was trained on ImageNet-1k with CodeFormer degradation, and IRControlNet was trained on filtered laion2b-en. For detailed training hyperparameters, please refer to the corresponding paper. They fine-tune SwinIR using the FFHQ dataset.

\begin{figure}[t]
    \centering
    \includegraphics[width=1\linewidth]{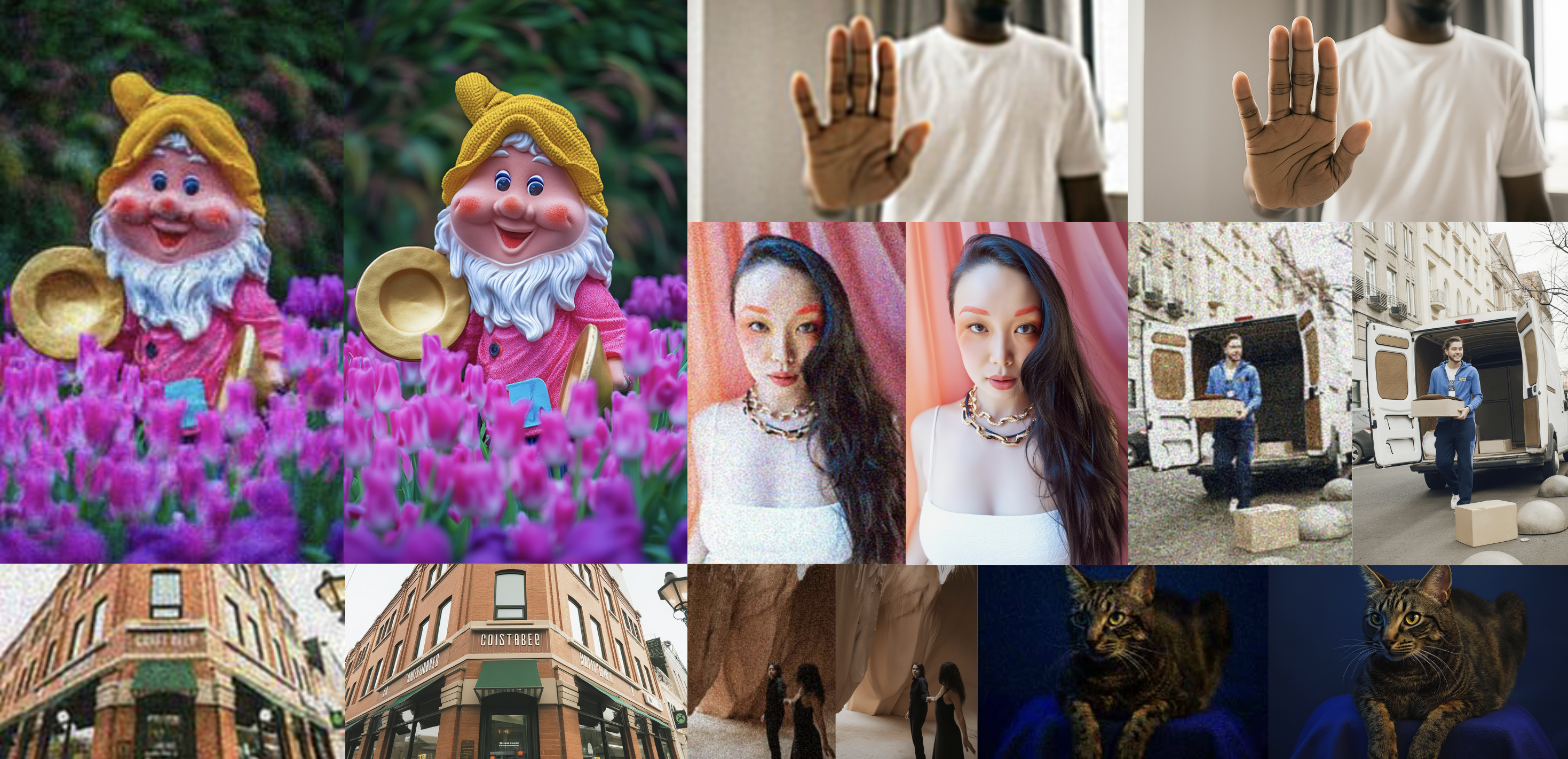} 
    \caption{\textbf{Team IPCV}}
    \label{fig:example}
    \vspace{-4mm}
\end{figure}

\subsection{IPCV}
\noindent\textbf{Description.} The Flux.1-dev-Controlnet-Upscaler is a deep learning model designed for image upscaling, integrating ControlNet with the Flux.1-dev architecture. It is based on a 12-billion-parameter rectified flow transformer, originally designed for high-quality image generation. The incorporation of ControlNet allows for guided upscaling by conditioning on additional inputs like depth maps, ensuring better structure preservation. The training process involved synthetic data degradation techniques, including Gaussian and Poisson noise addition, image blurring, and JPEG compression, making it robust for real-world image restoration. The model can be used with the diffusers library to enhance low-resolution images, and the implementation details are available on its Hugging Face page \cite{Flux1DevControlnetUpscaler}, shown in Fig.~\ref{fig:example}.
He adjusted the 4x scaling to 1x to maintain the original dimensions while enhancing pixel quality. 

\noindent\textbf{Implementation Details.} No training, only used inference. 

\section{Teams and Affiliations}
\label{sec:app_teams}
\subsection*{NTIRE 2025 team}
\noindent\textit{\textbf{Title: }} NTIRE 2025 Real-world Face Restoration Challenge\\
\noindent\textit{\textbf{Members: }} \\
Zheng Chen$^1$ (\href{mailto:zhengchen.cse@gmail.com}{zhengchen.cse@gmail.com}),\\
Jingkai Wang$^1$ (\href{mailto:jingkaiwang100@gmail.com}{jingkaiwang100@gmail.com}),\\
Kai Liu$^1$ (\href{mailto:normal.kliu@gmail.com}{normal.kliu@gmail.com}),\\
Jue Gong$^1$ (\href{mailto:g1017325431@gmail.com}{g1017325431@gmail.com}),\\
Lei Sun$^2$ (\href{mailto:leosun0331@gmail.com}{leosun0331@gmail.com}),\\
Zongwei Wu$^3$ (\href{zongwei.wu@uni-wuerzburg.de}{zongwei.wu@uni-wuerzburg.de}),\\
Radu Timofte$^3$ (\href{mailto:radu.timofte@uni-wuerzburg.de}{radu.timofte@uni-wuerzburg.de}),\\
Yulun Zhang$^1$ (\href{mailto:yulun100@gmail.com}{yulun100@gmail.com})\\
\noindent\textit{\textbf{Affiliations: }}\\
$^1$ Shanghai Jiao Tong University, China\\
$^2$ INSAIT, Bulgaria\\
$^3$ University of W\"urzburg, Germany\\

\section*{AllForFace}

\noindent\textit{\textbf{Title: }}Using Divide-and-Conquer for Blind Face Restoration

\noindent\textit{\textbf{Members: }} \\
Jianxing Zhang\textsuperscript{1}(\href{mailto:jx2018.zhang@samsung.com}{jx2018.zhang@samsung.com}), Jinlong Wu\textsuperscript{1}, Jun Wang\textsuperscript{1}, Zheng Xie\textsuperscript{1}, Hakjae Jeon\textsuperscript{2}, Suejin Han\textsuperscript{2}, Hyung-Ju Chun\textsuperscript{2}, Hyunhee Park\textsuperscript{2}

\noindent\textit{\textbf{Affiliations: }}\\
\textsuperscript{1}Samsung R\&D Institute China - Beijing (SRC-B)\\
\textsuperscript{2}Samsung MX(Mobile eXperience) Business

\section*{IIL}

\noindent\textit{\textbf{Title: }}Blind Face Restoration with One-Step Diffusion Framework

\noindent\textit{\textbf{Members: }} \\
Zhicun Yin\textsuperscript{1}(\href{mailto:cszcyin@outlook.com}{cszcyin@outlook.com}), Junjie Chen\textsuperscript{1}, Ming Liu\textsuperscript{1}, Xiaoming Li\textsuperscript{1}, Chao Zhou\textsuperscript{2}, Wangmeng Zuo\textsuperscript{1}

\noindent\textit{\textbf{Affiliations: }}\\
\textsuperscript{1}Harbin Institute of Technology\\
\textsuperscript{2}Shanghai Transsion Co, Ltd

\section*{PISA-MAP}

\noindent\textit{\textbf{Title: }}PiSA-MAP

\noindent\textit{\textbf{Members: }} \\
Weixia Zhang\textsuperscript{1}(\href{mailto:zwx8981@sjtu.edu.cn}{zwx8981@sjtu.edu.cn}), Dingquan Li\textsuperscript{2}, Kede Ma\textsuperscript{3}

\noindent\textit{\textbf{Affiliations: }}\\
\textsuperscript{1}Shanghai Jiao Tong University\\
\textsuperscript{2}Pengcheng Laboratory\\
\textsuperscript{3}City University of Hong Kong

\section*{MiPortrait}

\noindent\textit{\textbf{Title: }}MPSR

\noindent\textit{\textbf{Members: }} \\
Yun Zhang\textsuperscript{1}(\href{mailto:zhangyun9@xiaomi.com}{zhangyun9@xiaomi.com}), Zhuofan Zheng\textsuperscript{1}, Yuyue Liu\textsuperscript{1}, Shizhen Tang\textsuperscript{1}, Zihao Zhang\textsuperscript{1}, Yi Ning\textsuperscript{1}, Hao Jiang\textsuperscript{1}

\noindent\textit{\textbf{Affiliations: }}\\
\textsuperscript{1}Xiaomi Inc.

\section*{AIIA}

\noindent\textit{\textbf{Title: }}A Face Image Restoration Method Applying Pre-trained Models and Test-time Adaptation

\noindent\textit{\textbf{Members: }} \\
Wenjie An\textsuperscript{1}(\href{mailto:anwenjie1213@163.com}{anwenjie1213@163.com}), Kangmeng Yu\textsuperscript{1}

\noindent\textit{\textbf{Affiliations: }}\\
\textsuperscript{1}Harbin University of Technology

\section*{UpHorse}

\noindent\textit{\textbf{Title: }}DSS: Implicit Representation-Based Face Restoration with Diffusion Prior

\noindent\textit{\textbf{Members: }} \\
Yingfu Zhang\textsuperscript{1}(\href{mailto:zmund0717@gmail.com}{zmund0717@gmail.com}), Gang He\textsuperscript{1}, Siqi Wang\textsuperscript{1}, Kepeng Xu\textsuperscript{1}, Zhenyang Liu\textsuperscript{1}

\noindent\textit{\textbf{Affiliations: }}\\
\textsuperscript{1}Xidian University

\section*{CX}

\noindent\textit{\textbf{Title: }}CX

\noindent\textit{\textbf{Members: }} \\
Changxin Zhou\textsuperscript{1}(\href{mailto:changxin.zhou@bst.ai}{changxin.zhou@bst.ai}), Shanlan Shen\textsuperscript{1}, Yubo Duan\textsuperscript{1}

\noindent\textit{\textbf{Affiliations: }}\\
\textsuperscript{1}Black Sesame Technologies (Singapore) Pte Ltd

\section*{AIIALab}

\noindent\textit{\textbf{Title: }}DCMoE-RWFR

\noindent\textit{\textbf{Members: }} \\
Yiang Chen\textsuperscript{1}(\href{mailto:xantares606@gmail.com}{xantares606@gmail.com}), Kui Jiang\textsuperscript{1}, Jin Guo\textsuperscript{1}, Mengru Yang\textsuperscript{1}, Junjun Jiang\textsuperscript{1}

\noindent\textit{\textbf{Affiliations: }}\\
\textsuperscript{1}Harbin Institute of Technology

\section*{ACVLab}

\noindent\textit{\textbf{Title: }}ACVLab

\noindent\textit{\textbf{Members: }} \\
Jen-Wei Lee\textsuperscript{1}(\href{mailto:jemmy112322@gmail.com}{jemmy112322@gmail.com}), Chia-Ming Lee\textsuperscript{1}, Chih-Chung Hsu\textsuperscript{1,2}

\noindent\textit{\textbf{Affiliations: }}\\
\textsuperscript{1}Institute of Data Science, National Cheng Kung University\\
\textsuperscript{2}Institute of Intelligent Systems, National Yang Ming Chiao Tung University

\section*{HH}

\noindent\textit{\textbf{Title: }}HH

\noindent\textit{\textbf{Members: }} \\
Hu Peng\textsuperscript{1}(\href{mailto:hup22@mails.tsinghua.edu.cn}{hup22@mails.tsinghua.edu.cn}), Chunming He\textsuperscript{2}

\noindent\textit{\textbf{Affiliations: }}\\
\textsuperscript{1}Shenzhen International Graduate School, Tsinghua University\\
\textsuperscript{2}Duke University

\section*{Fustar-fsr}

\noindent\textit{\textbf{Title: }}Incremental Learning-Face Restoration Model

\noindent\textit{\textbf{Members: }} \\
Tingyi Mei\textsuperscript{1}(\href{mailto:18084795694@163.com}{18084795694@163.com}), Qizhao Lin\textsuperscript{1}, Jialiang Chen\textsuperscript{1}

\noindent\textit{\textbf{Affiliations: }}\\
\textsuperscript{1}Fujian Normal University

\section*{Night Watch}

\noindent\textit{\textbf{Title: }}Night Watch

\noindent\textit{\textbf{Members: }} \\
Kepeng Xu\textsuperscript{1}(\href{mailto:kepengxu11@gmail.com}{kepengxu11@gmail.com}), Siqi Wang\textsuperscript{1}, Yingfu Zhang\textsuperscript{1}, Zhenyang Liu\textsuperscript{1}, Gang He\textsuperscript{1}

\noindent\textit{\textbf{Affiliations: }}\\
\textsuperscript{1}Xidian University

\section*{IPCV}

\noindent\textit{\textbf{Title: }}Fluxoration

\noindent\textit{\textbf{Members: }} \\
Jameer Babu Pinjari\textsuperscript{1}(\href{mailto:jameer.jb@gmail.com}{jameer.jb@gmail.com})

\noindent\textit{\textbf{Affiliations: }}\\
\textsuperscript{1}Independent Researcher

{\small
\bibliographystyle{ieeenat_fullname}
\bibliography{main}
}